\documentclass{article}

\usepackage[table,dvipsnames]{xcolor}
\usepackage{microtype}
\usepackage{graphicx}
\usepackage{subfigure}
\usepackage{booktabs} % for professional tables

\usepackage{booktabs}         % for \toprule, \midrule, \bottomrule

\usepackage{amsmath}          % for math symbols
\usepackage{siunitx}          % optional for numeric alignment
\usepackage{colortbl}

\usepackage{comment}
\usepackage{mdframed}
\usepackage{makecell} 
\usepackage[normalem]{ulem}
\useunder{\uline}{\ul}{}

\usepackage{mathbbol}
\usepackage{hyperref}

\usepackage[accepted]{icml2025}

\usepackage{amsmath}
\usepackage{amssymb}
\usepackage{mathtools}
\usepackage{amsthm}

\usepackage[capitalize,noabbrev]{cleveref}

\usepackage{multicol}

\crefname{figure}{Figure}{Figures}
\crefname{table}{Table}{Tables}
\crefname{equation}{Eq.}{Equations}
\crefname{algorithm}{Algorithm}{Algorithms}
\crefname{section}{Section}{Sections}

\theoremstyle{plain}
\newtheorem{theorem}{Theorem}[section]
\newtheorem{proposition}[theorem]{Proposition}

\newtheorem{corollary}[theorem]{Corollary}
\theoremstyle{definition}
\newtheorem{definition}[theorem]{Definition}

\theoremstyle{remark}

\usepackage[textsize=tiny]{todonotes}

\mdfdefinestyle{highlightedBox}{
    linecolor=gray!20,          % Border color
    backgroundcolor=gray!5,   % Background color
    innertopmargin=5pt,       % Space before the start of the content
    innerbottommargin=2pt,    % Space after the end of the content
    roundcorner=4pt            % Round corner radius
    innerleftmargin=0pt,
    innerrightmargin=3pt,
    rightmargin=-0.2cm,
    leftmargin=-0.2cm
}

\usepackage{amsmath,amsfonts,bm}
\def\1{\bm{1}}

\def\dd{{\mathrm{d}}}
\DeclareMathAlphabet{\mathsfit}{\encodingdefault}{\sfdefault}{m}{sl}
\SetMathAlphabet{\mathsfit}{bold}{\encodingdefault}{\sfdefault}{bx}{n}

\def\gL{{\gL}}

\def\gZ{{\mathcal{Z}}}

\def\sP{{\mathbb{P}}}
\def\sQ{{\mathbb{Q}}}

\def\sm{{\mathbb{m}}}

\def\deltat{{\Delta t}}
\def\cat{{\mathrm{Cat}}}

\newcommand{\E}{\mathbb{E}}

\newcommand{\R}{\mathbb{R}}

\def\P{{\mathbb{P}}}

\def\Q{{\mathbb{Q}}}

\def\R{{\mathbb{R}}}

\definecolor{pearDark}{HTML}{2980B9}

\icmltitlerunning{Debiasing Guidance for Discrete Diffusion with Sequential Monte Carlo}

\begin{document}

\twocolumn[
\icmltitle{Debiasing Guidance for Discrete Diffusion with Sequential Monte Carlo}



\icmlsetsymbol{equal}{*}

\begin{icmlauthorlist}
\icmlauthor{Lee Cheuk-Kit}{equal,Cambridge}
\icmlauthor{Paul Jeha}{equal,DTU}
\icmlauthor{Jes Frellsen}{DTU}
\icmlauthor{Pietro Lio}{Cambridge}
\icmlauthor{Michael Albergo}{Harvard,IAIFI}
\icmlauthor{Francisco Vargas}{Cambridge}
\end{icmlauthorlist}

\icmlaffiliation{Cambridge}{University of Cambridge}
\icmlaffiliation{DTU}{Technical University of Denmark}
\icmlaffiliation{Harvard}{Society of Fellows, Harvard University}
\icmlaffiliation{IAIFI}{Institute for Artificial Intelligence and Fundamental Interactions, Massachusetts Institute of Technology}

\icmlcorrespondingauthor{Lee Cheuk-Kit}{brianlee.lck@gmail.com}

\icmlkeywords{Machine Learning, ICML}

\vskip 0.3in
]



\printAffiliationsAndNotice{\icmlEqualContribution} % otherwise use the standard text.

\begin{abstract}
Discrete diffusion models are a class of generative models that produce samples from an approximated data distribution within a discrete state space. Often, there is a need to target specific regions of the data distribution. Current guidance methods aim to sample from a distribution with mass proportional to $p_0(x_0) p(\zeta|x_0)^\alpha$ but fail to achieve this in practice. We introduce a Sequential Monte Carlo algorithm that generates unbiasedly from this target distribution, utilising the learnt unconditional and guided process. We validate our approach on low-dimensional distributions, controlled images and text generations. For text generation, our method provides strong control while maintaining low perplexity compared to guidance-based approaches.

\end{abstract}
\begin{figure}[h!]
    \centering
    \includegraphics[width=1\linewidth]{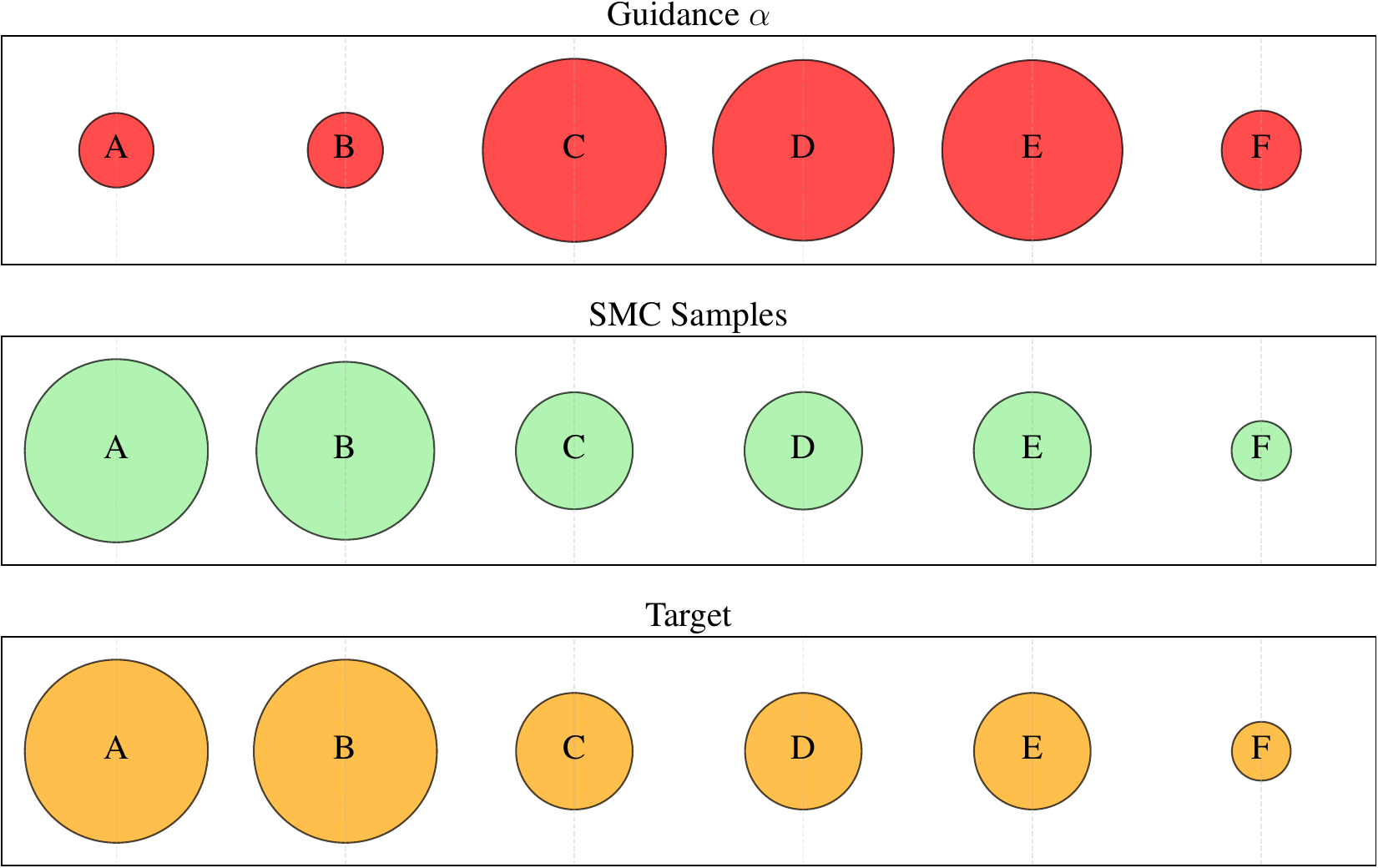}
     \vspace{-1.5em} 
    \caption{Visualization of a 1D discrete distribution with six states (A-F). The size of each disk represents probability mass. Orange circles show the target tempered distribution, green circles represent our SMC-based method which accurately approximates the target, while red circles show standard guidance failing to match the target distribution.}
    \label{fig:guided-process-does-not-hit}
\end{figure}

\section{Introduction} \label{sec:introduction}

Discrete Diffusion models generate approximate samples from a data distribution $p_0(x_0)$ by gradually evolving samples from a simple base distribution $p_1(x_1)$ through a Continuous-Time Markov Chain (CTMC) \citep{diffusion-jiaxin, campbell2022continuoustimeframeworkdiscrete, sedd}. While these models learn unconditional distributions, practical applications such as graph generation \citep{vignac2023digress}, protein co-design \citep{multimodal-flow} or text generation \citep{sedd} require controlled generation. For a conditioning variable $\zeta$ and temperature parameter $\alpha$, we aim to sample from a tempered conditional distribution proportional to $p_0(x_0) p(\zeta|x_0)^\alpha$. When $\alpha = 1$, this recovers the conditional $p_0(x_0 | \zeta)$. Higher values ($\alpha > 1$) biases sampling toward the conditioning signal $\zeta$, while lower values ($\alpha < 1$) promotes diversity. To sample from this tempered distribution, guidance methods modify the unconditional diffusion transition rates \cite{unlock_guidance}. However, guidance fails to sample from the intended distribution \cite{what-does-guidance-do, bradley2024classifierfreeguidancepredictorcorrector}, producing corrupted samples at high $\alpha$ values \cite{what-does-guidance-do}. Recent works propose finetuning approaches using reinforcement learning to sample from tempered distributions \citep{domingoenrich2025adjointmatchingfinetuningflow, venkatraman2024amortizing, NEURIPS2023_fc65fab8, black2024trainingdiffusionmodelsreinforcement, clark2024directlyfinetuningdiffusionmodels, uehara2025inferencetimealignmentdiffusionmodels}.
% , these methods require access to $p_0(\zeta|x_0)^\alpha$, which we assume not available in our setting.\\
In this work, we address the theoretical and practical challenges of sampling from tempered distributions in discrete diffusion models. This work makes the following contributions:
\begin{enumerate}
\item We derive the exact transition rates required to sample from the tempered distribution with unnormalised mass $p_0(x_0) p(\zeta|x_0)^\alpha$ in discrete diffusion.
\item We propose an algorithm based on Sequential Monte Carlo, that asymptotically samples from the intended tempered distribution by exploiting previously forgone properties of the guided transition rate matrix, without any additional learning.
\item We validate our approach through low-dimensional experiments and demonstrate its effectiveness on controlled image and text generation, achieving strong conditional control with low perplexity on text dataset.
% \item We validate our approach through low-dimensional experiments and demonstrate its practicality in controlled image and text generation. Notably on text dataset, our method shows strong conditional control while maintaining low perplexity.
\end{enumerate}

% Related works such as \citep{domingoenrich2025adjointmatchingfinetuningflow, venkatraman2024amortizing} can also be utilised to finetune models to sample the tempered distribution using a reinforcement learning like setup, assuming access to $p_0(\zeta|x_0)^\alpha$ which we do not.   

\section{Background Work}
\subsection{Continuous Time Markov chains}
A Continuous Time Markov Chain (CTMC) $\{X_t\}_{t\in [0, 1]}$ is a Markov process on a finite state space $\mathcal{X} = \{1, \dots, S\}$. The evolution of $X_t$ is governed by a time-dependent rate matrix $R_t : \mathcal{X} \times \mathcal{X} \to \mathbb{R}$, which defines infinitesimal transition probabilities $p_{t+\Delta t | t}(x_{t+\Delta t} | x_t)$ given by
\begin{align}
\delta_{x_{t+\Delta t}, x_t} + R_t(x_t, x_{t+\Delta t}) \Delta t + o(\Delta t),
\end{align}
where $\delta_{a, b}$ is $1$ if $a = b$ and $0$ otherwise.

The probability mass $p_t$ of $X_t$ evolves according to the Kolmogorov Forward Equation (KFE):
\begin{align}
\partial_t p_t(x) = \underbrace{\sum_{y \neq x} p_t(y) R_t(y, x)}_\text{incoming mass} - \underbrace{\sum_{y \neq x} p_t(x) R_t(x, y)}_\text{outgoing mass},
\end{align}
which can be written in vector form as:
\begin{align} \label{eq:eq-kolmogorov-forward}
\partial_t p_t = R_t^\intercal p_t,
\end{align}
where $R_t$ satisfies mass conservation:
\begin{align}
R_t(x, x) = -\sum_{y \neq x} R_t(x, y).
\end{align}
For simplicity, we define,
\begin{align}
    R_t(x) = \sum_{y \neq x} R_t(x, y)
\end{align}
To obtain approximate samples $\tilde{X}_t$, one may use the Euler sampling algorithm \citep{multimodal-flow}, initialising $\tilde{X}_0 \sim p_0$ and updating subsequent samples in intervals of $\Delta t$ following:  
\begin{align}
\tilde{X}_{t+\Delta t} &\sim \tilde{p}^\text{Euler}_{t+\deltat|t}(\cdot | \tilde{x}_t), \\ \tilde{p}^\text{Euler}_{t+\deltat|t}(\tilde{x}_{t+\deltat} | \tilde{x}_t) &\propto \delta_{\tilde{x}_{t+\deltat}, \tilde{x}_t} + R_t(\tilde{x}_t, \tilde{x}_{t+\deltat}) \Delta t.
\end{align}
A more comprehensive overview including the treatment of reverse-time CTMC can be found in \Cref{app:primer-ctmc}.
\subsection{Discrete Diffusion Model} \label{sec:diffusion}
Discrete diffusion models are generative models that rely on a pair of stochastic processes: a forward process corrupting data distribution $p_0$ into base distribution $p_1$, and its learned reverse process reconstructing $p_0$ from $p_1$. Discrete diffusion models formulate these as Continuous Time Markov Chains on finite state spaces $\mathcal{X}$ with $|\mathcal{X}| = S$. Discrete diffusion models require learning the reverse process.  \citet{campbell2022continuoustimeframeworkdiscrete} learn the reverse rate matrix through likelihood objectives, while \citet{sedd} instead learn the reverse rate matrix using a score entropy objective. In their work, they propose the masking diffusion model, which extends the state space $\mathcal{X}$ with a masking state $\sm = S+1$ to form an extended state space $\overline{\mathcal{X}} = \{1, \dots, S, \sm\}$. At time $t$, the masking process sends the current state to the masking state $\sm$ according to a time-dependent noise schedule $\sigma: [0, 1] \rightarrow \R^+$, with transition probability $p^\text{mask}_{t|0}$:
\begin{align}
p^\text{mask}_{t|0}(x_t|x_0) = \delta_{x_t, x_0} e^{-\sigma(t)} + \delta_{x_t, \sm}(1 - e^{-\sigma(t)})
\end{align}
For sufficiently large $\sigma(1)$, the masking process mixes to a point mass at $\sm$ at time $t=1$. To sample from a masking diffusion model, we start at the masking state $\sm$ at $t=1$ and apply Euler sampling in reverse time with learned rate matrix $\tilde{R_t}$. For high-dimensional data like text, the process extends to $\mathcal{X}^d$ by corrupting each dimension independently \citep{diffusion-jiaxin}.

\section{Guidance} \label{sec:cond-guide}
Let $\{X_t\}_{t \in [0,1]}$ be a reverse-time CTMC with probability mass $p_t$ and rate matrix $R_t$. For a conditioning variable $\zeta$, we write $p_t(\zeta| x_t)$ and $p_t(x_t| \zeta)$ as the conditioned probability desnity of $\zeta$ given $X_t = x_t$ and the conditioned probability mass of $x_t$ given $\zeta$ respectively.

With guidance scale $\alpha$, we aim to sample from the tempered distribution $p_0(x_0)p_0(\zeta | x_0)^\alpha / \gZ_\alpha$, where $\gZ_\alpha$ is a normalising constant. When $\alpha = 1$, this recovers $p_0(x_0 | \zeta)$. Higher values ($\alpha > 1$) bias sampling toward high likelihood regions of $\zeta$, while lower values promote diversity.

Guidance modifies $R_t$ to $R_t^\alpha$ (\Cref{def:R_alpha}) under the premise that it samples correctly from the intended tempered distribution. However, guidance fails to do so when $\alpha \neq 1$ as shown in \Cref{fig:guided-process-does-not-hit}.

\begin{definition}
\label{def:R_alpha}
The guided rate matrix $R_t^\alpha$ is defined as,
\begin{align}
    \forall x \neq y. \; R_t^\alpha(x, y | \zeta) &=  R_t(x, y) \left[\frac{p_t(\zeta | y)}{p_t(\zeta | x)}\right]^\alpha, \\ 
    \forall x. \; R_t^\alpha(x, x | \zeta) &= -R_t^\alpha(x|\zeta)
\end{align}
\end{definition} 
We defer the learning of guided rate matrix to \Cref{sec:learn-guidance}. One might expect the guided rate matrix $R_t^\alpha$ to represent the time-reversal corrupting process starting at the tempered distribution $p_0(x_0) p_0(\zeta | x_0)^\alpha / \gZ_\alpha$. However, as shown in \Cref{prop:true-tempered}, this is not the case.
\begin{mdframed}[style=highlightedBox]
\begin{proposition} \label{prop:true-tempered}
Let $M_t[\cdot]$ denote the evolution of probability mass at time $t$ under the corrupting process $p^{\text{corrupt}}_{t|0}$ of $X_t$, the unconditional diffusion model. Define $p_t^{\alpha,\text{true}}$ as:
\begin{align*}
p_t^{\alpha,\text{true}} = M_t\big[p_0(\cdot) \, p_0(\zeta | \cdot)^\alpha / \gZ_\alpha\big].
\end{align*}
Then,
\begin{align*}
p_t^{\alpha,\text{true}}(x_t) \propto p_t(x_t) \mathbb{E}\big[p_0(\zeta | X_0)^\alpha | X_t = x_t\big].
\end{align*}

The true tempered rate matrix $R_t^{\alpha,\text{true}}$ given by 
\begin{align*}
\forall x \neq y, \; R_t^{\alpha,\text{true}}(x, y | \zeta) &= R_t(x, y) \tfrac{\mathbb{E}[p_0(\zeta | X_0)^\alpha | X_t = y]}{\mathbb{E}[p_0(\zeta | X_0)^\alpha | X_t = x]}, \\ 
\forall x, \; R_t^{\alpha,\text{true}}(x, x | \zeta) &= -\sum_{y \neq x} R_t^{\alpha,\text{true}}(x, y | \zeta),
\end{align*}
is rate matrix of time reversal of the corruption process starting from $p_0(\cdot)p_0(\zeta | \cdot)^\alpha / \gZ_\alpha$
\end{proposition}
\end{mdframed}
From \Cref{prop:true-tempered}, it follows that $R_t^\alpha = R_t^{\alpha,\text{true}}$ holds only for $\alpha = 1$. Furthermore, for $\alpha = 1$, guidance requires that the base distribution $p_1$ is independent of the conditioning variable $\zeta$ to correctly sample from the conditioned distribution $p_0(X_0 | \zeta)$, as shown in \Cref{lemma:correct-guidance-alpha-1}.
\begin{corollary} \label{lemma:correct-guidance-alpha-1}
If $p_1(x_1 | \zeta) = p_1(x_1)$, then guidance samples correctly from the conditioned distribution $p_0(x_0 | \zeta)$. 
\end{corollary}
\vspace{-0.7em}
While \citep{unlock_guidance} presents similar results to \Cref{lemma:correct-guidance-alpha-1}, our lemma establishes that the independence condition is necessary.\\
Learning $R_t^{\alpha, true}$ most often requires expensive simulation-based objectives to estimate the gradients of a reverse path-wise KL divergence \citep{,domingoenrich2025adjointmatchingfinetuningflow,DEFT,uehara2025inferencetimealignmentdiffusionmodels}. Therefore, we focus instead on sampling from $p_t^{\alpha}(x_t) \propto p_t(x) p_t(\zeta|x_t)^\alpha$.

% Unfortunately, there is no cheap simulation-free way of learning $R_t^{\alpha, true}$ as most approaches require a simulation-based objective \citep{,domingoenrich2025adjointmatchingfinetuningflow,DEFT,uehara2025inferencetimealignmentdiffusionmodels} that estimates the gradients of a reverse path-wise KL divergence. Thus, instead of sampling from $p_t^{\alpha,\text{true}}$ in what follows, we will shift the focus to sampling from $p_t^{\alpha}(x_t) = p_t(x) p_t(\zeta|x_t)^\alpha$.
\vspace{-0.2em}
\section{Debiasing Guidance with Sequential Monte Carlo } %\label{sec:debiasing-guidance}

Our objective is to sample from the tempered distribution $\frac{1}{Z_0^{\alpha}}p_0(x_0)p_0(\zeta |x_0)^{\alpha}$. To that end, we first introduce an importance sampling method, which we later leverage through resampling. 

\subsection{Importance Sampling the Tempered Distribution}
We consider the importance sampling method to compute the expectation of a function $h$ under the distribution of probability mass $p_t^\alpha(x_t) = \frac{1}{\gZ_t^{\alpha}}p_t(x_t)p_t(\zeta|x_t)^\alpha$ where $\gZ_t^{\alpha}$ is the normalising constant. We write the expectation as:
\begin{align}
    \E_{x \sim p_t^\alpha}[h(x)] =\frac{1}{\gZ_t^\alpha} \sum_{x \in \mathcal{X}} h(x) p_t(x) p_t(\zeta|x_t)^\alpha,
\end{align}
For importance sampling, we consider a proposal in the form of a reverse time CTMC $\{Y_t\}_{t \in [0,1]}$, with rate matrix $Q_t$ and initial distribution $p_1$. We construct unnormalised importance weight $\{W_t\}_{t \in [0,1]}$ such that:
\begin{align}
    \E_{x \sim p_t^\alpha}[h(x)] = \frac{\E[W_t \cdot h(Y_t)]}{\E[W_t]}
\end{align}
where $\E$ is an expectation over the joint distribution $\text{Law}(W_t, Y_t)$.

\Cref{prop:cont-resampling} provides one form of the unnormalised importance weights $\{W_t\}_{t \in [0,1]}$.

\begin{mdframed}[style=highlightedBox]
\begin{proposition} \label{prop:cont-resampling}
Let proposal $\{Y_t\}_{t\in[0,1]}$ be a reverse-time CTMC with rate matrix $Q_t$ and initial distribution $p_1(\cdot)$. Define $\{W_t\}_{t\in[0,1]}$ by 
\begin{align*}
W_t = \frac{\mathrm{d} \mathbb{P}_t^{\mathrm{base}}}{\mathrm{d}  \mathbb{Q}_t} \left(\frac{\mathrm{d} \mathbb{P}^{ \alpha=1}_t }{\mathrm{d}\mathbb{P}_t^{\mathrm{base}} } \right)^\alpha, %\!\!\!\!(Y_t),
\end{align*}
where we have path space measures $\sP_t^{\text{base}}=\mathrm{Law}{\{X_\tau\}_{\tau\in[t,1]}}$, $\sP^{\alpha=1}_t=\mathrm{Law}{\{X_\tau\mid\zeta\}_{\tau\in[t,1]}}$ and  $\sQ_t = \mathrm{Law}{\{Y_\tau\}_{\tau\in[t,1]}}$.

Then the Radon-Nikodym derivatives can be written as,
\begin{align*}
\ln \frac{\mathrm{d} \mathbb{P}_t}{\mathrm{d} \mathbb{Q}_t}^{\!\mathrm{ base}} 
 =  &\displaystyle\sum_{\substack{ \tau  \geq t\\ Y_{\tau^+}\neq Y_\tau}} 
\!\!\ln R_t(Y_{\tau^+}, Y_\tau) 
\\&+ \int_{1}^t  Q_\tau(Y_t) - R_\tau(Y_t) \, \mathrm{d} \tau
\end{align*}
and
\begin{align*}
\ln\! \frac{\mathrm{d} \mathbb{P}_t^{ \alpha=1} }{\mathrm{d}\mathbb{P}_t^{\mathrm{base}} } 
\!= &\!\!\!\!\!\displaystyle\sum_{\substack{ \tau  \geq t\\ Y_{\tau^+}\neq Y_\tau}} 
\!\!\!\! \ln R_t^{\alpha=1}\!(Y_{\tau^+}, Y_\tau|\zeta)
 \\
&+\!\!\int_{1}^t  \!\!\!R_\tau(Y_\tau) - R^{\alpha=1}_\tau(Y_\tau|\zeta) \mathrm{d} \tau,
\end{align*}
such that for any $p_t^{\alpha}$-integrable function $h$ and time $t \in [0, 1]$, we have
\begin{align*}
    \E_{x \sim p_t^\alpha}[h(x)] = \frac{\E[W_t \cdot h(Y_t)]}{\E[W_t]},
\end{align*}
where the expectation is taken over $\text{Law}(W_t, Y_t)$
\end{proposition}
\end{mdframed}
A proof of the proposition can be found in \Cref{sec:proof-cont-resampling}. Due to the modular nature of our result, it seamlessly extends to stochastic differential equations for more details please see \Cref{sec:sde_prop}.

For practical implementation, we approximate the CTMC by discretising time into $T$ steps:  $1 = t_1 > t_2 > \dots > t_T = 0$. For times $t < s$, we denote the transitions derived from rate matrices $Q_t$, $R_t$, and $R^{\alpha=1}_t$ as $q_{t|s}(x_t | x_s)$, $p_{t|s}(x_t | x_s)$, and $p_{t|s}(x_t | x_s, \zeta)$ respectively. To this end, we consider a discretisation of \Cref{prop:cont-resampling} in \Cref{sec:discretised-weights}. We also present \Cref{alg:the-algorithm} in \Cref{sec:resampling-smc}, a pseudo-code of the discretised version of the proposed sampling method.

\subsection{Resampling and Sequential Monte Carlo}
\label{sec:resampling-smc}
To approximate sample from the tempered distribution $p_0^\alpha$, we leverage resampling. Given a set of $K$ weights and samples $(w_t^{(i)}, y_t^{(i)}) \sim \text{Law}(W_t, Y_t)$, resampling allows us to obtain approximate samples from $p_t^\alpha$ by sampling the Categorical distribution,
\begin{align}
\cat\left(\left\{\frac{w_t^{(i)}}{\sum_{j=1}^K w_t^{(j)}}\right\}_{i=1}^K, y_t^{(i)}\right),
\end{align}
where $y_t^{(i)}$ is sampled with probability $\frac{w_t^{(i)}}{\sum_{j=1}^K w_t^{(j)}}$.

This suggests an algorithm to obtain samples from the tempered distribution $p_0(x_0)p_0(\zeta|x_0)^\alpha$ by resampling at time $t=0$. However, this approach suffers from a significant challenge: weight degeneracy. As $t$ approaches 0, the variance of weights $W_t$ increases dramatically, requiring an impractical number of samples.\\
Sequential Monte Carlo (SMC) addresses this challenge by performing resampling steps at intermediate times $0 < t < 1$, obtaining approximate samples of $p_t^\alpha$ while resetting the weights to maintain the importance sampling equality. Effectively, this eliminates sample from low likelihood regions and duplicates those in high likelihood regions.\\
The resampling step is flexible in both algorithm choice and timing. In \Cref{sec:pixel-level}, we explore partial resampling where only a subset of samples participate. Resampling is typically triggered when the effective sample size (ESS) falls below a threshold, where ESS is defined as:
\begin{align}
\text{ESS}_t^K = \left(\sum_{i=1}^K w^{(i)}_t\right)^2 / \sum_{i=1}^K (w^{(i)}_t)^2 \in [1, K]
\end{align}

We emphasize that the resampling procedure allows us to leverage importance sampling to generate asymptotically unbiased samples rather than purely computing expectations. The full algorithm to sample from the tempered distribution can be found in \cref{alg:the-algorithm} detailed in Algo. \ref{algo:main-algo} (App. \ref{sec:discretised-main-algorithm}) for our experiments.

\begin{algorithm}[h]
  \caption{Main Algorithm}
  \label{alg:the-algorithm}
  \begin{algorithmic}[1]
    \REQUIRE Number of particles $K$; Proposal Rate Matrix Proposal Transition $Q_t$; Unconditional rate matrix $R_t$; Time grid $\{t_l\}_{l=1}^T$ for potential resampling; ESS threshold $\text{ESS\_THRESHOLD}$; Resampling algorithm \texttt{resample}.
    \STATE \textbf{Initialisation:}
    \STATE Sample $\{y_1^{(i)}\}_{i=1}^K$ i.i.d. from $p_1$.
    \STATE Set $\hat{w}_1^{(i)} \leftarrow 1$ for $i=1,\dots,K$.
    \FOR{$l = 1$ to $T$}
      \FOR{$i = 1$ to $K$}
        \STATE \textcolor{gray!80}{\COMMENT{Step 1: Evolve samples and weights}}
         \STATE Obtain $(y_{l+1}^{(i)}, \hat{w}_{l+1}^{(i)})$ by evolving joint system $(Y_t, W_t)$ from $(y_{l}^{(i)}, \hat{w}_{l}^{(i)})$ at time $t_l$
         \vspace{0.5em}
      \ENDFOR
      \STATE \textcolor{gray!80}{\COMMENT{Step 3: Resample; see \Cref{sec:resample} for other resampling algorithms}} 
      \STATE Set $\text{ESS} \leftarrow \left(\sum_{i=1}^K 
      \hat{w}^{(i)}_{l}\right)^2/ \sum_{i=1}^K (\hat{w}^{(i)}_{l})^2$
      \vspace{0.5em}
      \IF{$\text{ESS} \leq \text{ESS\_THRESHOLD}$}
      \vspace{0.5em}
          \STATE Set $y_{l+1}^{(i)}, \hat{w}_{l+1}^{(i)} \leftarrow \texttt{resample}(\{y_{l}^{(i)}\}, \{\hat{w}_l^{(i)}\})$ 
      \ENDIF
    \ENDFOR
    \STATE \textbf{Output:} 
    \STATE Particles $\{x_T^{(i)}\}_{i=1}^K$
  \end{algorithmic}
\end{algorithm}

\section{Numerical Verification} \label{sec:numerical}
We validate our algorithm's ability to sample from the tempered distribution $p_0(x_0) p_0(\zeta | x_0)^\alpha / \gZ_\alpha$ using low-dimensional examples. These experiments use explicitly specified probability mass $p_0(x_0)$ and conditional likelihood $p_0(\zeta|x_0)$, enabling direct computation of the guided ratio matrix. This setup verifies algorithmic correctness independent of learning effects and allows evaluation of the KL divergence between sampled and target distributions.
\begin{figure}[h!]
    \centering
    \includegraphics[width=\linewidth]
    {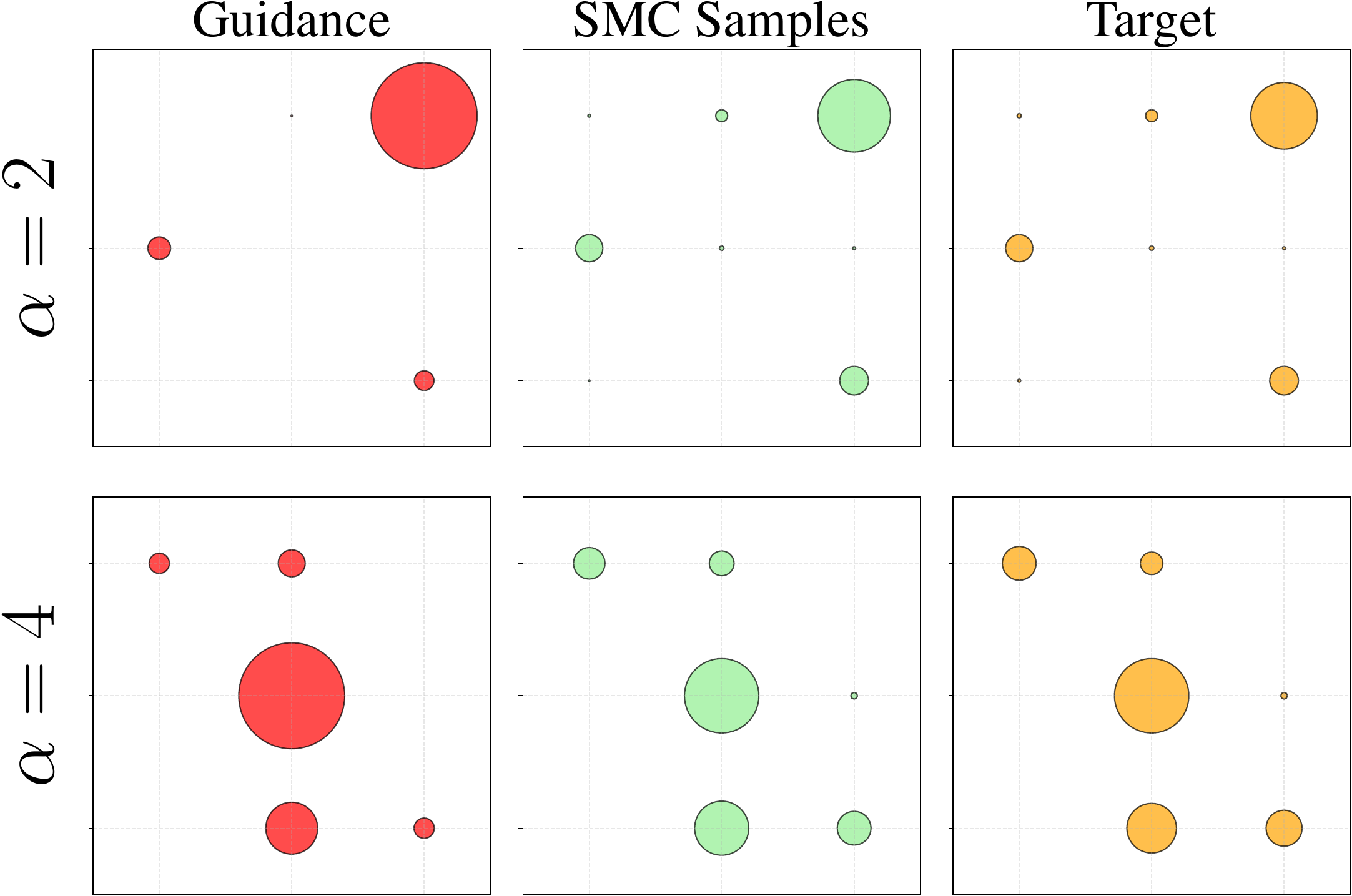}
    \vspace{-2em}
    \caption{A two dimensional discrete distribution with nine states. The target distribution is in orange, standard guidance is in red and our SMC based method in green Top panel shows $\alpha=2$, bottom panel $\alpha=4$. In both cases, guidance-based sampling deviates from the target distribution, while SMC samples are significantly closer.}
    \label{fig:toy-examples}
\end{figure}

\paragraph{Experimental setup}

We evaluate on state spaces of dimension one ($\mathcal{V}$) and two ($\mathcal{V}^2$), where $\mathcal{V}$ is a finite vocabulary. Our proposal uses per-dimension Euler sampling \citep{multimodal-flow} with 100 discretization steps and 50,000 samples. For each dimension, we generate 30 different tempered target distributions $\pi \propto p_0(x_0)p_0(\zeta|x_0)^\alpha$ by combining different choices of $p_0(x_0)$, $p_0(\zeta|x_0)$, and temperature $\alpha$. We sample from each target using both guidance and SMC, then compute KL divergences between the true tempered distribution and the empirical distributions $\sigma^\text{SMC}$ and $\sigma_\text{Guided}$. Results in \Cref{tab:forward-KL} show that guided processes significantly deviate from targets while SMC achieves approximate sampling, demonstrating superior sampling across all combinations. Qualitative examples are shown in \Cref{fig:guided-process-does-not-hit,fig:toy-examples}.

% Given a vocabulary $\mathcal{V}$, we consider a one and two dimensional state space, $\mathcal{V}$ and $\mathcal{V}^2$ respectively. The proposal uses per-dimension Euler sampling \citep{multimodal-flow} with 100 discretisation steps and 50000 samples. To demonstrate guidance sampling failure, we compare guided CTMC proposals with and without resampling steps. For each dimension, we generate 30 combinations of $p_0(x_0)$, $p_0(\zeta | x_0)$, and temperature $\alpha$, computing KL divergences between the tempered target $\pi \propto p_0(x_0) p_0(\zeta | x_0)^\alpha$ and both empirical distributions $\sigma^\text{SMC}$ and $\sigma_\text{Guided}$. Qualitative results in \Cref{fig:guided-process-does-not-hit,fig:toy-examples} show guided processes deviating significantly from targets while  SMC achieves approximate target sampling, with KL divergences in \Cref{tab:forward-KL} confirming superior sampling across all combinations.

\begin{table}[h!t]
\centering
\begin{tabular}{c|c|cccc}
\toprule
\textbf{Dimension} & \textbf{$|\mathcal{V}|$} & $\text{KL}(\pi  |  |  \sigma_{\text{SMC}})$ & $\text{KL}(\pi  |  |  \sigma_{\text{Guided}})$ \\
\toprule
$1$ & 50 &\cellcolor{blue!15}{$0.002 \pm 0.001$} & $0.004 \pm 0.003$       \\
$2$ & 3 &\cellcolor{blue!15}{$0.035 \pm 0.179$} & $0.074\pm 0.082$       \\
$2$ & 10 &\cellcolor{blue!15}{$0.013 \pm 0.012$} & $0.081 \pm 0.052$       \\ 
$2$ & 100 &\cellcolor{blue!15}{$1.725 \pm 0.743$} & $6.690 \pm 1.194$      \\
$2$ & 200 &\cellcolor{blue!15}{$5.948 \pm 1.042$} & $12.906 \pm 0.542$      \\
\bottomrule
\end{tabular}
\caption{Forward KL divergence between the target tempered distribution $\pi$ and the empirical distribution of samples from the guided process and  SMC algorithm. Method with a lower KL divergence is highlighted.}
\label{tab:forward-KL}
\end{table}

\section{Experiments}
\Cref{sec:numerical} demonstrated SMC's ability to sample tempered distributions with sufficient particles and discretisation steps. In practice, for image and text generation tasks, computing rate matrices is computationally intensive, limiting feasible particle counts and step sizes. For these practical settings, we use the guided rate matrix as our proposal in SMC, with guidance temperature $\beta$ distinct from the SMC temperature $\alpha$, and provide guidance on resampling strategies.

\subsection{Learning the Guidance Term} \label{sec:learn-guidance}

We consider two approaches to obtain the guided rate matrix $R_t^\alpha$, \textbf{DEFT} and \textbf{CFG}.

\paragraph{DEFT} For discrete diffusion models, we extend Doob's h-transform Efficient FineTuning (DEFT, \citet{DEFT}) by showing that the conditional rate matrix $R_t^{\alpha=1}$ decomposes into an unconditional rate matrix $R_t$ and a guidance term $p_t(\zeta|y) / p_t(\zeta|x)$: $R_t^{\alpha=1}(x, y)=R_t(x,y)\times p_t(\zeta|y) / p_t(\zeta|x)$. We model the guidance term with a neural network $g$ and parameterize the conditional rate matrix as $R_t^{\alpha=1}(x, y | \zeta) = R_t(x, y) g(x, y, t)$ while keeping the unconditional rate parameters fixed. We use this guidance term in both pixel level image generation and controlled text generation.

\paragraph{CFG} In text generation, we extend classifier-free guidance \citep{classifier-free-guidance}. We obtain the conditional rate matrix $R_t^{\alpha=1}(x, y|\zeta)$ by finetuning the learned unconditional rate matrix $R_t(x, y|\zeta)$,  concatenating conditions to the inputs. For $\alpha > 1$, the guided rate matrix follows $\left[p_t(\zeta|y) / p_t(\zeta|x)\right]^\alpha = \left[R_t^{\alpha=1}(x, y | \zeta) / R_t(x, y) \right]^\alpha$

There are other ways of finding a guided rate matrix. \citet{unlock_guidance, vignac2023digress} consider a first-order Taylor approximation of the guidance term; \citet{training-free-guidance, li2024derivativefreeguidancecontinuousdiscrete} consider a training-free approach by Monte Carlo estimates of the guidance term. 

In what follows we present results where we adopted the CFG approach and refer the reader to \Cref{app:deft_result} for results with DEFT. We first discuss partial resampling, a key component of our SMC implementation, and then detail our experiments and findings.

\subsection{Partial Resampling} \label{sec:partial-resampling}
SMC methods suffer from mode collapse in high dimensions when using few particles. In this scenario, most particle weights decay rapidly, leaving only a small subset with significant weights. During resampling, particles with low weights are discarded, often resulting in only one or two unique particles. \Cref{fig:class-conditional-image-generation} (middle) demonstrates this effect: a batch of 16 particles collapses into identical images.

We adopt a partial resampling scheme which effectively mitigates mode collapse by resampling only a subset of particles. We select the $K$ most and least weighted samples for resampling, which preserves sample diversity and maintains unbiased sampling from the target distribution as shown in \Cref{fig:class-conditional-image-generation} (right). We detail partial resampling algorithms in  \Cref{sec:resample}. In our experiments, we use \Cref{algo:partial_resampling}, set $K = \lfloor N/4\rfloor$, resampling half of all generated samples.

% \subsection{Choosing the Proposal}

% SMC algorithms require effective proposal selection. We use the guided rate matrix as our proposal, with guidance temperature $\beta$ distinct from the SMC temperature $\alpha$.

% Choosing a good proposal is crucial in obtaining samples for SMC algorithms. In our experiments, we consider using the guided rate matrix as the proposal. As the guided rate matrix warrants a different temperature parameter from the SMC algorithm, we denote a guidance temperature parameter with $\beta$, and a SMC temperature parameter $\alpha$.

\subsection{Class-Conditional Pixel-Level Image Generation} \label{sec:pixel-level}
We evaluate our method on MNIST class-conditional generation using $28 \times 28$ grayscale images with pixel values in $\{0,...,255\}$. We first detail the experimental setup; we then study partial resampling for mitigating mode collapse in SMC and analyze its impact on generation control.

\subsubsection{Dataset and Training Details}
We train an unconditional SEDD model \citep{sedd} parameterized by a U-Net \citep{ronneberger2015unetconvolutionalnetworksbiomedical} for 20k steps. Using 20\% of MNIST, we finetune a guidance network for 4k steps with DEFT using the SEDD denoising score entropy loss. The guidance network is a U-Net with learned condition embeddings added to each layer, at 50\% parameters of the unconditional network.

\begin{figure}[h!]
    \centering
    \includegraphics[width=\linewidth]{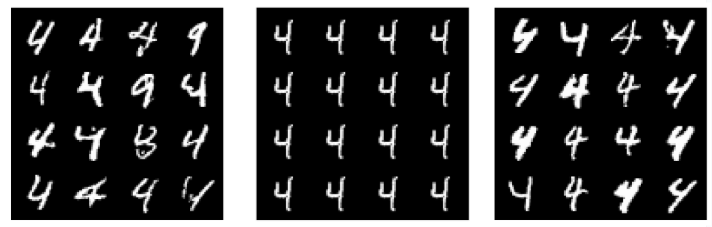}
    \caption{\textbf{Class-Conditioned Image Generation $\zeta = 4$}: From left to right: Independent guidance samples,  SMC samples without partial resampling, and  SMC samples with partial resampling.  SMC with partial resampling generates high-quality samples of digit 4 while avoiding mode collapse.}
    \label{fig:class-conditional-image-generation}
\end{figure}
% \vspace{-0.9em}
\subsubsection{Does Partial Resampling mitigate Mode Collapse?}
The MNIST dataset visually demonstrates mode collapse and the effectiveness of partial resampling (\Cref{sec:partial-resampling}). Using SMC with guidance temperature $\beta=1$ and SMC temperature $\alpha=3$, we observe mode collapse in \Cref{fig:class-conditional-image-generation} (center), where 16 particles converge to identical images. Partial resampling (\Cref{fig:class-conditional-image-generation}, right) maintains both diversity and stronger conditional control compared to naive guidance (\Cref{fig:class-conditional-image-generation}, left), which occasionally generates incorrect digits.

% The visual nature of the MNIST dataset allows us to demonstrate the mode collapse problem and the effectiveness of the partial resampling approach mentioned in \Cref{sec:partial-resampling} more concretely. We adopt a SMC algorithm with guidance temperature $\beta=1$ and SMC temperature $\alpha=3$.

% The center image of \Cref{fig:class-conditional-image-generation} provides clear evidence of mode collapse: a batch of 16 particles collapses into identical images. The rightmost image of \Cref{fig:class-conditional-image-generation} shows images obtained using an partial resampling scheme. It can be clearly seen that the images are able to maintain stronger conditional control compared to the naive guidance which generates incorrect digits at times, while maintaining a high diversity. 

\subsubsection{Does SMC improve conditional control?}
\begin{figure}[htbp]
    \centering
    \includegraphics[width=\linewidth]{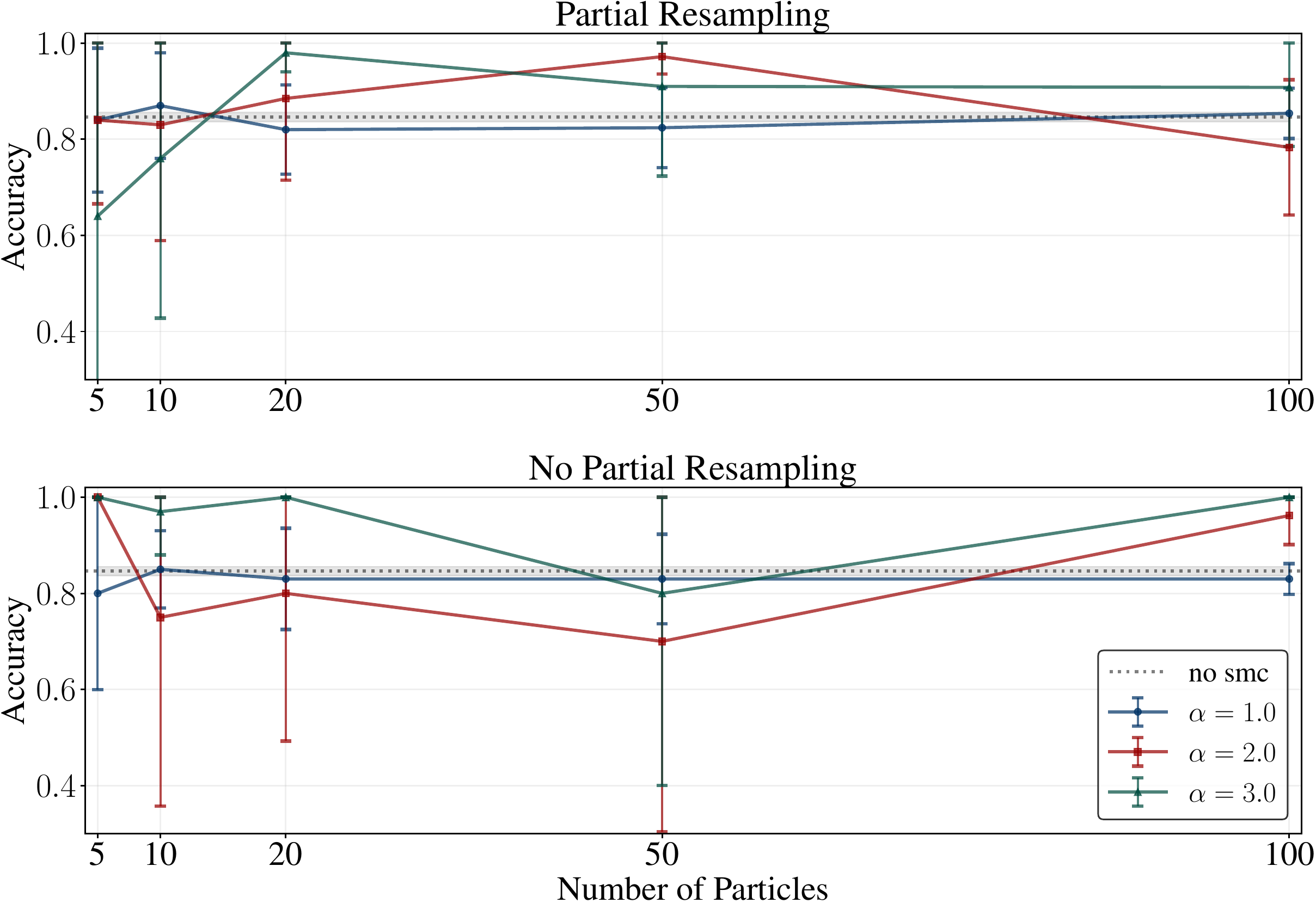}
    \vspace{-1em}
    \caption{\textbf{Accuracy on Class-Conditioned Image generation.}  SMC classification accuracy versus particle count for different twist temperatures ($\alpha$), with and without partial resampling. Partial resampling and higher $\alpha$ achieve better peak accuracy. Dotted lines show baseline accuracy without SMC. Values averaged over 50 runs.}
    \label{fig:class-conditioned-mnist-metrics}
\end{figure}

% We adopt a partial resampling effectively mitigates mode collapse by resampling only a subset of particles. We select the $K$ most and least weighted samples for resampling, which preserves sample diversity and maintains unbiased sampling from the target distribution as shown in \Cref{sec:resample}. In our experiments, we set $K = \lfloor N/4\rfloor$, resampling half of all generated samples.
For class-conditioned generation, we evaluate sample consistency with target conditions using a pre-trained digit classifier. \Cref{fig:class-conditioned-mnist-metrics} shows accuracy for SMC temperatures $\alpha \in \{1, 2, 3\}$ and $N$ number of particle in $\{5, 10, 20, 50, 100\}$, using $R_t^{\alpha=1}$ as the proposal. We compute accuracies within each SMC run and confidence intervals across 10 independent runs. Higher SMC temperatures yield better accuracy across all particle counts. Partial resampling achieves higher accuracy with lower variance, which indicates that it effectively reduces mode collapse.
% For class-conditioned generation, we compute an accuracy metric, measuring whether the samples is consistent with the supplied condition according to a pre-trained digit classifier. In \Cref{fig:class-conditioned-mnist-metrics}, we report accuracy with SMC temperature $\alpha = \{1, 2, 3\}$ using the conditioned process $R_t^{\alpha=1}$ as the proposal. The number of particles in $\{5, 10, 20, 50, 100\}$. An accuracy is computed across samples in the same SMC run while a confidence interval is computed across 10 independent runs. As expected, it can be seen that with a higher SMC temperature generally leads to higher accuracy for different number of particles. Partial resampling also achieves higher accuracy but with lower variance, likely suggesting that less effect of mode collapse.

Experiments on MNIST demonstrate that our method successfully addresses two key challenges in conditional generation. Partial resampling effectively mitigates mode collapse while maintaining sample diversity, and higher SMC temperatures improve conditional control across all particle counts. These results validate our theoretical framework on a standard image generation task.

\subsection{Controlled Text Generation}
We validate the approach on three controlled text generation: Sentiment Controlled Generation,  Toxicity Controlled Generation, and Text Infilling. These high-dimensional experiments demonstrates the algorithm practicality.

\subsubsection{Dataset and Training Details}
In all experiments, we use the pre-trained 320M non-embedding parameters \texttt{sedd-medium}, a Score Entropy Discrete Diffusion (SEDD) model \citep{sedd}. When guiding with CFG we finetune  \texttt{sedd-medium} and when guiding with DEFT we finetune \texttt{sedd-small} for 2k steps to serve as the guidance term $p_t(\zeta|y)/p_t(\zeta|x)$  for the large \texttt{sedd-medium} model, on task-specific datasets:

\paragraph{Toxicity Controlled Generation} Following \citep{dexperts}, we use human-annotated comments from the Jigsaw Unintended Bias in Toxicity Classification challenge\footnote{\url{https://bit.ly/3cvG5py}}, with texts labeled toxic when 50\% or more annotators mark them as such. A prompt \texttt{The generated text is toxic} is concatenated to the transformer input as the condition.

\paragraph{Sentiment Controlled Generation.} We follow the setup in \citep{structured-vornoi-sampling}, using the Stanford Sentiment Treebank, a dataset of movie reviews, as the finetuning dataset. A prompt \texttt{The generated text is of positive sentiment} or \texttt{The generated text is of negative sentiment} is concatenated to the transformer input as the condition.

\paragraph{Text Infilling.} We finetune on 10\% of the OpenWebText training dataset. For each sentence, randomly sampled indices (\texttt{START}, \texttt{END}) define the token range to mask. The masked sentence $\zeta$ is concatenated to the model input throughout generation. The entire partial masked sentence \texttt{[PREFIX]} \texttt{[MASK]} \texttt{[SUFFIX]} is passed to the network as the condition.

We evaluate samples using both fluency and conditional control metrics, averaging across 5 runs and 50 particles for SMC or 250 particles for guidance, with 100 discretisation steps and we employ partial resampling. For fluency, we compute generative perplexity \citep{dexperts} using \texttt{GPT2-XL} as the reference model, which quantifies how likely a sentence is to appear in the reference model. For sentiment and toxicity tasks, we use a LoRA-finetuned \texttt{GPT2-XL} to account for stylistic differences in the finetuning dataset. For conditional metrics, we use a pre-trained classifier for sentiment accuracy, Perspective API for toxicity scores, and measure how many unmasked tokens are successfully generated.

\subsubsection{Ablating Guidance and SMC temperature}
We study how guidance temperature $\beta$ and SMC temperature $\alpha$ impact both conditional control and fluency metrics across toxicity, sentiment, and infilling tasks.
\vspace{-1em}
\paragraph{Our findings} SMC-based methods improve considerably on conditional control metrics for low-to-mid guidance temperature $\beta$ and with high SMC temperature $\alpha$. For high guidance temperature, the guidance baseline improves conditional metrics but at the cost of higher perplexity. This implies that guidance-generated samples drift from the distribution of the fine-tuning dataset. Hence, we conclude that the optimal approach is to combine SMC with high SMC temperature and mid-to-low guidance temperature.
% With a temperature parameter for both SMC and guidance, one might be interested in the question: \textit{How do the collected samples varies as we vary the two parameters?}

% To measure the quality of our samples, we adopt a fluency metric and a task-specific performance metric measuring the effectiveness of control. In all SMC experiments, we compute metrics by averaging 5 independent runs with 50 particles. For guidnace, we sample 250 particles. 100 discretisation steps are used for both setups.

% For fluency, we adopt generative perplexity following prior work \citep{dexperts}. Generative perplexity is a likelihood metric that tracks how likely a sentence is to appear in a reference model. For infilling, the reference model is chosen to be \texttt{GPT2-XL}. For sentiment, and toxicity, we notice that the FineTuning dataset generally has low likelihood under \texttt{GPT2-XL} due to the stylistic difference (sentence length/vocabulary). Thus, we adopt a \texttt{GPT2-XL} model finetuned using LoRA, as the reference model.

% For performance, we adopt an accuracy metric for sentiment controlled generation as computed by a pretrained classifier, a toxicity score for toxicity controlled generation computed using the Perplexity API, and an accuracy metric for infilling measuring how many blanks are being filled.

\begin{figure}[h!]
    \centering
    \includegraphics[width=\linewidth]{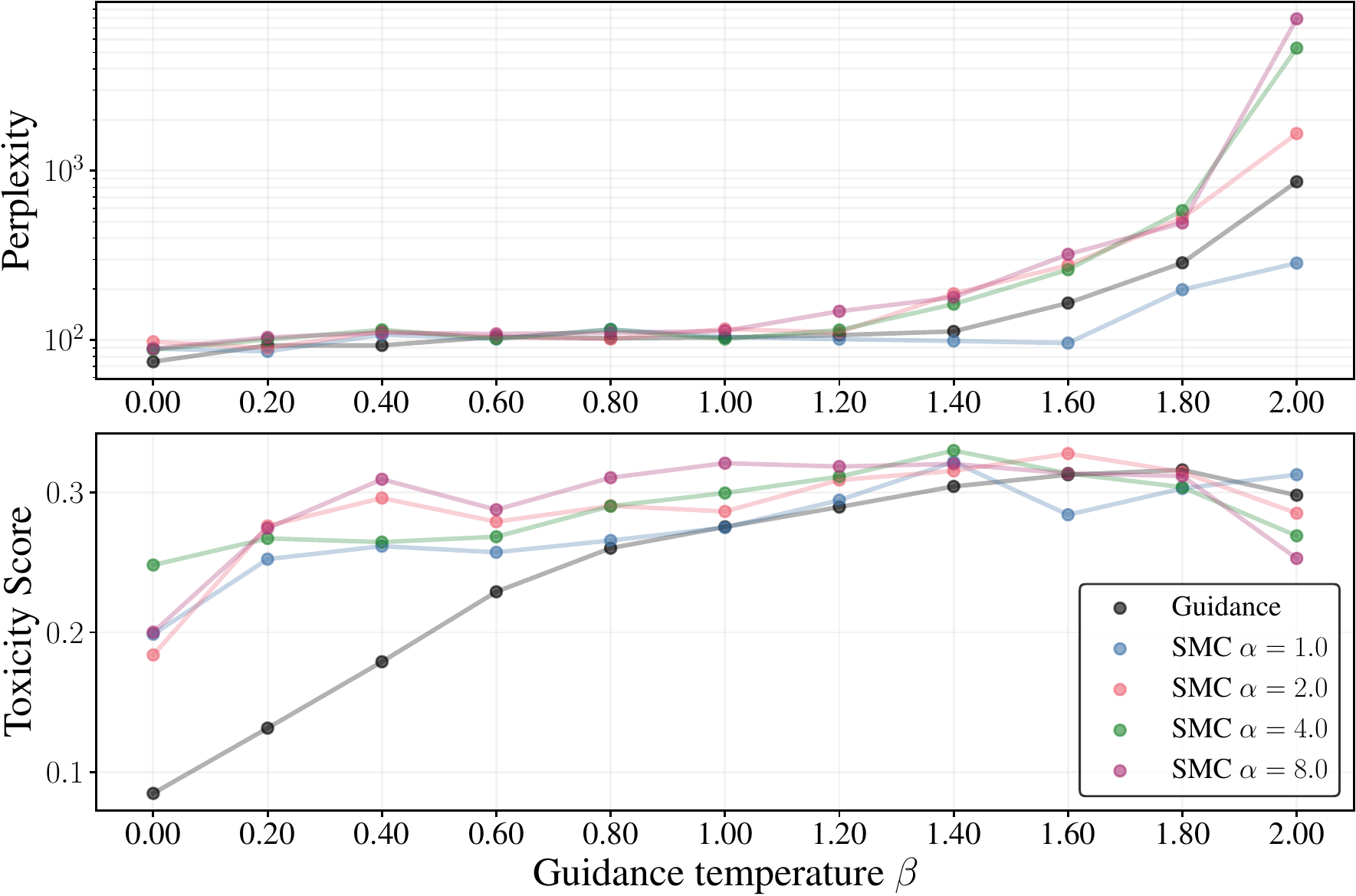}
    \vspace{-1em}
    \caption{\textbf{Toxicity controlled generation:} SMC methods show improved toxicity scores across all guidance temperatures $\beta$, demonstrating enhanced generation control. While baseline performance matches SMC at high $\beta$ values, the perplexity deteriorates significantly.}
    \label{fig:toxic-cfg}
\end{figure}

\begin{figure}[h!]
    \centering
    \includegraphics[width=\linewidth]{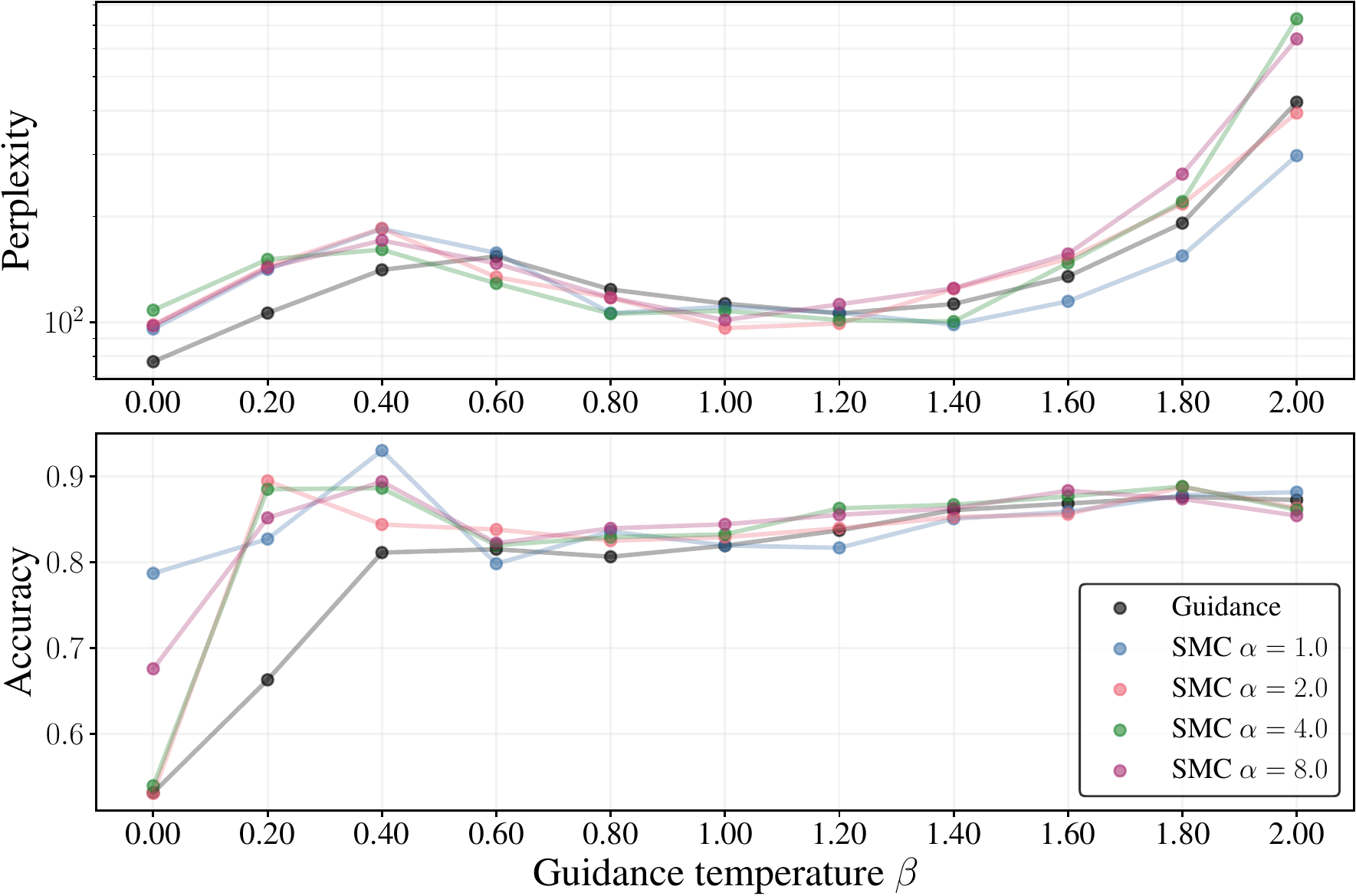}
    \vspace{-1em}
    \caption{\textbf{Sentiment control generation:} For guidance temperatures $\beta \in \{0.2, 0.4\}$, SMC approaches achieve notably better accuracy compared to guidance, though at the cost of slightly higher perplexity. For guidance temperature greater than $1.4$ and for $\alpha=1$, SMC method achieves lower perplexity compared to guidance.}
    \label{fig:sentiment-cfg}
\end{figure}

\begin{figure}[h!]
    \centering
    \includegraphics[width=1\linewidth]{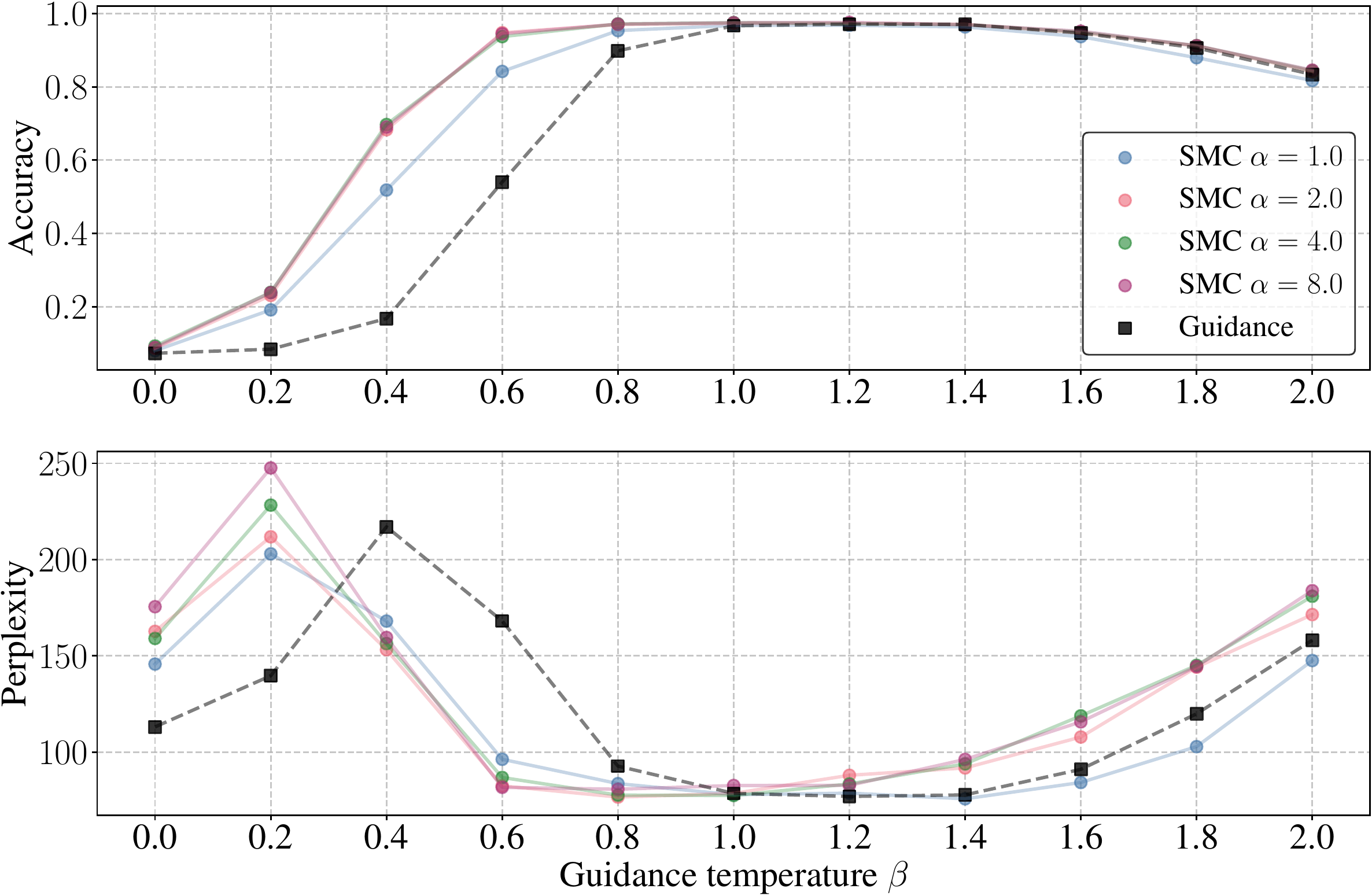}
     \vspace{-1.5em} 
    \caption{\textbf{Text infilling}: SMC method shows improved accuracy scores accross guidance temperature. In addition for $\beta \in \{0.4, 0.6\}$ SMC methods show in addition improved perplexity, as well as for $\beta \geq 1.6$ and $\alpha = 1$.}
    \vspace{-1em}
    \label{fig:cfg-infill}
\end{figure}

In more detail, \Cref{fig:toxic-cfg}, \Cref{fig:sentiment-cfg}, and \Cref{fig:cfg-infill} show the generative perplexity and performance metrics across guidance temperature $\beta=\{0.0, 0.2, 0.4, 0.6, 0.8, 1.0, 1.2, 1.4, 1.6, 1.8, 2.0\}$ and SMC temperature $\alpha=\{1, 2, 4, 8\}$. Across all three tasks, SMC methods with high SMC temperature $\alpha$ consistently outperform the guidance method. For high guidance temperature $\beta$ the gap between SMC and the guidance method reduces. However, perplexity deteriorates at high guidance temperature which indicates that the proposal is unable to sample from high likelihood region of the target tempered distribution.

For toxicity control, SMC consistently achieves better toxicity scores until $\beta > 1.4$, while maintaining comparable perplexity. In sentiment control, SMC shows improved accuracy until $\beta > 1.2$, with slightly higher perplexity. At $\beta > 1.4$ and $\alpha=1$, SMC achieves lower perplexity than standard guidance while maintaining the same accuracy as the other methods. For text infilling, SMC shows superior or on-par accuracy across all guidance temperatures and better perplexity for $\beta \in \{0.4, 0.6\}$; or at $\alpha =1$ and $\beta > 1.4$ SMC shows reduced perplexity while maintaining the same accuracy.
As $\beta$ increases beyond 1.4, we observe a common pattern across all tasks: the performance gap between SMC and guidance narrows, but at the cost of significant perplexity deterioration, particularly for higher $\beta$ values. 
% In all cases, it can be seen that using SMC with a high temperature consistently outperforms the guidance method. At a guidance temperature, the gap becomes smaller, however, perplexity usually deteriorates, signifying the proposal is unable to sample from high likelihood region of the target tempered distribution.
\begin{table*}[htbp]
\centering
\small
\setlength{\tabcolsep}{3pt}
\renewcommand{\arraystretch}{1.2}
\resizebox{\linewidth}{!}{
\begin{tabular}{l l | c c c c c c c c | c c | c}
\toprule
\multicolumn{2}{l|}{\textbf{Method}} 
& \multicolumn{8}{c|}{\textbf{Sequential Monte Carlo}} 
& \multicolumn{2}{c|}{\textbf{Guidance}} 
& \textbf{Prompting} \\
\midrule
% -- First header line: SMC Scale
& \multicolumn{1}{l|}{SMC Temp. $\alpha$}
& \multicolumn{2}{c|}{1.0} 
& \multicolumn{2}{c|}{2.0} 
& \multicolumn{2}{c|}{4.0} 
& \multicolumn{2}{c|}{8.0} 
& \multicolumn{2}{c|}{N/A} 
& N/A \\
\toprule
% -- Second header line: Guidance temperature (centered)
& \multicolumn{1}{l|}{Guidance Temp. $\beta$} 
& 1.0 & \multicolumn{1}{c|}{1.2} 
& 1.0 & \multicolumn{1}{c|}{1.2} 
& 1.0 & \multicolumn{1}{c|}{1.2} 
& 1.0 & \multicolumn{1}{c|}{1.2} 
& 1.0 & \multicolumn{1}{c|}{1.2} 
& N/A \\
\midrule
\multicolumn{2}{l|}{\textbf{Perplexity} }
& $84.410$\scriptsize{$\pm8.647$} 
& \multicolumn{1}{l|}{$93.225$\scriptsize{$\pm7.281$}}
& $80.500$\scriptsize{$\pm3.554$} 
& \multicolumn{1}{l|}{$83.767$\scriptsize{$\pm7.457$}}
& $77.538$\scriptsize{$\pm11.209$} 
& \multicolumn{1}{l|}{$85.170$\scriptsize{$\pm7.253$}}
& $80.283$\scriptsize{$\pm6.825$} 
& \multicolumn{1}{l|}{$78.388$\scriptsize{$\pm6.606$}}
& $80.283$\scriptsize{$\pm6.825$} 
& \multicolumn{1}{l|}{$78.388$\scriptsize{$\pm6.606$}}
& $44.673$\scriptsize{$\pm11.687$} \\

\multicolumn{2}{l|}{\textbf{Accuracy} $\uparrow$}
& $0.839$\scriptsize{$\pm0.047$} 
& \multicolumn{1}{l|}{$0.840$\scriptsize{$\pm0.065$}}
& \cellcolor{blue!15}$0.882$\scriptsize{$\pm0.048$} 
& \multicolumn{1}{l|}{$0.877$\scriptsize{$\pm0.050$}}
& $0.875$\scriptsize{$\pm0.048$} 
& \multicolumn{1}{l|}{$0.872$\scriptsize{$\pm0.057$}}
& $0.868$\scriptsize{$\pm0.059$} 
& \multicolumn{1}{l|}{$0.875$\scriptsize{$\pm0.060$}}
& $0.837$\scriptsize{$\pm0.006$} 
& \multicolumn{1}{l|}{$0.837$\scriptsize{$\pm0.006$}}
& $0.548$\scriptsize{$\pm0.105$} \\
\bottomrule
\end{tabular}
}
\caption{Performance metrics for \textbf{sentiment control} using Sequential Monte Carlo, Guidance, and Prompting. 
\colorbox{blue!15}{Light blue} marks the best performing method.}
\label{tab:sentiment}
\end{table*}

%==========================================================
% 1) Table for TOXICITY control only (rows 1-2)
%============================================================
\begin{table*}[htbp]
\centering
\small
\setlength{\tabcolsep}{3pt}
\renewcommand{\arraystretch}{1.2}
\resizebox{\linewidth}{!}{
\begin{tabular}{l l | c c c c c c c c | c c | c}
\toprule
\multicolumn{2}{l|}{\textbf{Method}} 
& \multicolumn{8}{c|}{\textbf{Sequential Monte Carlo}} 
& \multicolumn{2}{c|}{\textbf{Guidance}} 
& \textbf{Prompting} \\
\midrule
% -- First header line: SMC Scale
& \multicolumn{1}{l|}{SMC Temp. $\alpha$}
& \multicolumn{2}{c|}{1.0} 
& \multicolumn{2}{c|}{2.0} 
& \multicolumn{2}{c|}{4.0} 
& \multicolumn{2}{c|}{8.0} 
& \multicolumn{2}{c|}{N/A} 
& N/A \\
\toprule
% -- Second header line: Guidance temperature (centered)
& \multicolumn{1}{l|}{Guidance Temp. $\beta$} 
& 1.0 & \multicolumn{1}{c|}{1.2} 
& 1.0 & \multicolumn{1}{c|}{1.2} 
& 1.0 & \multicolumn{1}{c|}{1.2} 
& 1.0 & \multicolumn{1}{c|}{1.2} 
& 1.0 & \multicolumn{1}{c|}{1.2} 
& N/A \\
\midrule
\multicolumn{2}{l|}{\textbf{Perplexity}}
& $66.558$\scriptsize{$\pm9.085$} 
& \multicolumn{1}{l|}{$58.913$\scriptsize{$\pm1.945$}}
& $62.192$\scriptsize{$\pm6.998$} 
& \multicolumn{1}{l|}{$88.650$\scriptsize{$\pm3.325$}}
& $89.125$\scriptsize{$\pm4.062$} 
& \multicolumn{1}{l|}{$109.200$\scriptsize{$\pm3.518$}}
& $77.000$\scriptsize{$\pm13.420$} 
& \multicolumn{1}{l|}{$88.800$\scriptsize{$\pm1.549$}}
& $67.475$\scriptsize{$\pm8.467$} 
& \multicolumn{1}{l|}{$70.525$\scriptsize{$\pm10.079$}}
& $44.319$\scriptsize{$\pm13.885$} \\

\multicolumn{2}{l|}{\textbf{Toxicity} $\uparrow$}
& $0.451$\scriptsize{$\pm0.244$} 
& \multicolumn{1}{l|}{$0.384$\scriptsize{$\pm0.231$}}
& $0.577$\scriptsize{$\pm0.232$} 
& \multicolumn{1}{l|}{$0.547$\scriptsize{$\pm0.222$}}
& $0.595$\scriptsize{$\pm0.231$} 
& \multicolumn{1}{l|}{\cellcolor{blue!15} $0.611$\scriptsize{$\pm0.208$}}
& $0.561$\scriptsize{$\pm0.252$} 
& \multicolumn{1}{l|}{$0.558$\scriptsize{$\pm0.230$}}
& $0.455$\scriptsize{$\pm0.244$} 
& \multicolumn{1}{l|}{$0.505$\scriptsize{$\pm0.246$}}
& $0.130$\scriptsize{$\pm0.123$} \\
\bottomrule
\end{tabular}
}
\caption{Performance metrics for \textbf{toxicity control} using Sequential Monte Carlo, Guidance, and Prompting. 
\colorbox{blue!15}{Light blue} marks the best performing method.}
\label{tab:toxicity}
\end{table*}

\begin{table*}[ht!]
\centering
\small
\setlength{\tabcolsep}{5pt}
\renewcommand{\arraystretch}{1.2}
\resizebox{\textwidth}{!}{
\begin{tabular}{l l | c c c c | c c | c}
\toprule
% -----------------------------------------------------------------
% Header row #1
% -----------------------------------------------------------------
\multicolumn{2}{l|}{\textbf{Method}}
& \multicolumn{4}{c|}{\textbf{Sequential Monte Carlo}}
& \multicolumn{2}{c|}{\textbf{Guidance}}
& \textbf{Reconstruction} \\
\midrule
% -----------------------------------------------------------------
% Header row #2: Twist Scale
% -----------------------------------------------------------------
 & \multicolumn{1}{l|}{SMC Temp. $\alpha$}
 & 1.0 & 2.0 & 4.0 & 8.0
 & \multicolumn{2}{c|}{N/A}
 & N/A \\
\toprule
% -----------------------------------------------------------------
% Header row #3: Guidance Temperature
% -----------------------------------------------------------------
 & \multicolumn{1}{l|}{Guidance Temp. $\beta$}
 & 1.0 & 1.0 & 1.0 & 1.0
 & \multicolumn{2}{c|}{1.0}
 & N/A \\
\midrule
% -----------------------------------------------------------------
% Data rows
% -----------------------------------------------------------------

\multicolumn{2}{l|}{\textbf{Perplexity}}
 & $65.048$\scriptsize{$\pm30.951$} & $67.062$\scriptsize{$\pm42.959$}
 & $73.562$\scriptsize{$\pm26.980$} & $80.583$\scriptsize{$\pm45.478$}
 & \multicolumn{2}{c|}{$63.915$\scriptsize{$\pm31.375$}}
 & $53.697$\scriptsize{$\pm26.543$} \\

\multicolumn{2}{l|}{\textbf{BERTScore} $\uparrow$}
 & $0.583$\scriptsize{$\pm0.038$} & $0.584$\scriptsize{$\pm0.032$}
 & \cellcolor{green!15}{$0.587$\scriptsize{$\pm0.029$}}& $0.576$\scriptsize{$\pm0.038$}
 & \multicolumn{2}{c|}{$0.583$\scriptsize{$\pm0.037$}}
 & \cellcolor{blue!15}{$0.595$\scriptsize{$\pm0.038$}} \\

\multicolumn{2}{l|}{\textbf{GLEU} $\uparrow$}
 & $0.701$\scriptsize{$\pm0.021$} & $0.703$\scriptsize{$\pm0.022$}
 & \cellcolor{blue!15}{$0.708$\scriptsize{$\pm0.019$}} & $0.699$\scriptsize{$\pm0.022$}
 & \multicolumn{2}{c|}{$0.701$\scriptsize{$\pm0.021$}}
 & $0.706$\scriptsize{$\pm0.021$} \\

\multicolumn{2}{l|}{\textbf{Accuracy} $\uparrow$}       & $0.992$\tiny{$\pm0.015$} & $0.996$\tiny{$\pm0.014$} & \cellcolor{green!15}{$0.998$\tiny{$\pm0.013$}} & \cellcolor{green!15}{$0.998$\tiny{$\pm0.015$}} & \multicolumn{2}{c|}{$0.991$\tiny{$\pm0.018$}} & \cellcolor{blue!15}{$1.000$\tiny{$\pm0.000$}}  \\
\bottomrule
\end{tabular}
}
\caption{Performance of guidance methods on text infilling: Sequential Monte Carlo 
(varying twist temperature $\alpha$) plus a row of guidance temperature 
(all set to 1.0), Guidance, and Reconstruction guidance. Values show 
mean $\pm$ standard deviation. $\uparrow$/$\downarrow$ indicate better 
performance for higher/lower values. \colorbox{blue!15}{Light blue} cell 
marks the best result. \colorbox{green!15}{Light green} cell 
marks the best result between Guidance and SMC.}
\label{tab:infill}
\end{table*}

\subsubsection{Scaling up Discretisation Step}
We further investigate sample qualities by increasing discretization steps to 1000, effectively allocating 10x compute.  For text infilling, we employ two additional metrics: BERTScore \cite{bert-score} with the DeBERTa model \cite{he2021debertadecodingenhancedbertdisentangled}, and GLEU-4 \cite{wu2016googlesneuralmachinetranslation}. For sentiment and toxicity controlled generation, we compare against a baseline prompting method, while for infilling, we evaluate against a reconstruction method that overwrites the sampling trajectory with the partially masked sentence. 

\Cref{tab:infill}, \Cref{tab:sentiment}, and \Cref{tab:toxicity} summarise our empirical results. SMC with a high SMC temperature significantly outperforms the base guidance method across all tasks. Notably, for toxicity and sentiment-controlled generation, our SMC-based method significantly outperforms a simple prompting strategy in generating strong conditions. However, for infilling and despite their theoretical guarantees, guidance and SMC approaches underperform compared to the reconstruction method, even though it is known to be biased \citep{wu2024practical}. This indicates an inadequate learning of the guided rate matrix. While SMC can strengthen the base guidance method, it cannot fully compensate for an imperfectly learned guided rate matrix.

\section{Conclusion and Future Work} \label{sec:conclusion}

This work addresses a fundamental challenge in diffusion models: accurate sampling from tempered distribution. We make three key contributions. First, we derive the exact transition rate to sample from the tempered distribution; second, we propose an SMC-based algorithm that asymptotically samples from the tempered distribution; third, we demonstrate its practicality in controlled generation. Our empirical results validate the practicality of our theoretically correct algorithm. In low dimensions, it significantly outperforms standard guidance methods. In high dimensions, our method achieves superior control while maintaining sample quality. However, our experiments also reveal that the effectiveness depends on the quality of the learned guided rate matrix. \\
Lastly, \Cref{prop:cont-resampling} transfers naturally to continuous diffusion models, which represents a promising avenue for future research. 

% In conclusion, the work introduced an algorithm to correct the bias in guidance for discrete diffusion model with theoretical guarantees. A thorough empirical investigation is carried out on different choices in the  SMC algorithm, in particular, the choice of resampling algorithm and the proposal distribution.

% \Cref{prop:cont-resampling} transfers naively to continuous diffusion models. While discrete diffusion models are chosen as the subject of study in this paper due to theoretical simplicity in the derivation, as future direction, we hope to explore the use of the SMC algorithm in the continuous case.

\section*{Acknowledgement}
The authors would like to thank Junhua Chen for extensive dicussions on the implementation of Sequential Monte Carlo Methods.

\section*{Impact Statement}
This paper presents work whose goal is to advance the field of 
Machine Learning. There are many potential societal consequences 
of our work, none which we feel must be specifically highlighted here.

\nocite{langley00}

\bibliography{main}

\begin{thebibliography}{49}
\providecommand{\natexlab}[1]{#1}
\providecommand{\url}[1]{\texttt{#1}}
\expandafter\ifx\csname urlstyle\endcsname\relax
  \providecommand{\doi}[1]{doi: #1}\else
  \providecommand{\doi}{doi: \begingroup \urlstyle{rm}\Url}\fi

\bibitem[Albergo \& Vanden-Eijnden(2023)Albergo and
  Vanden-Eijnden]{albergo2023buildingnormalizingflowsstochastic}
Albergo, M.~S. and Vanden-Eijnden, E.
\newblock Building normalizing flows with stochastic interpolants, 2023.
\newblock URL \url{https://arxiv.org/abs/2209.15571}.

\bibitem[Albergo \& Vanden-Eijnden(2024)Albergo and
  Vanden-Eijnden]{albergo2024nets}
Albergo, M.~S. and Vanden-Eijnden, E.
\newblock Nets: A non-equilibrium transport sampler.
\newblock \emph{arXiv preprint arXiv:2410.02711}, 2024.

\bibitem[Albergo et~al.(2023)Albergo, Boffi, and
  Vanden-Eijnden]{albergo2023stochasticinterpolantsunifyingframework}
Albergo, M.~S., Boffi, N.~M., and Vanden-Eijnden, E.
\newblock Stochastic interpolants: A unifying framework for flows and
  diffusions, 2023.
\newblock URL \url{https://arxiv.org/abs/2303.08797}.

\bibitem[Amini et~al.(2024)Amini, Du, and
  Cotterell]{structured-vornoi-sampling}
Amini, A., Du, L., and Cotterell, R.
\newblock Structured voronoi sampling, 2024.
\newblock URL \url{https://arxiv.org/abs/2306.03061}.

\bibitem[Black et~al.(2024)Black, Janner, Du, Kostrikov, and
  Levine]{black2024trainingdiffusionmodelsreinforcement}
Black, K., Janner, M., Du, Y., Kostrikov, I., and Levine, S.
\newblock Training diffusion models with reinforcement learning, 2024.
\newblock URL \url{https://arxiv.org/abs/2305.13301}.

\bibitem[Blattmann et~al.(2023)Blattmann, Dockhorn, Kulal, Mendelevitch,
  Kilian, Lorenz, Levi, English, Voleti, Letts, Jampani, and
  Rombach]{blattmann2023stablevideodiffusionscaling}
Blattmann, A., Dockhorn, T., Kulal, S., Mendelevitch, D., Kilian, M., Lorenz,
  D., Levi, Y., English, Z., Voleti, V., Letts, A., Jampani, V., and Rombach,
  R.
\newblock Stable video diffusion: Scaling latent video diffusion models to
  large datasets, 2023.
\newblock URL \url{https://arxiv.org/abs/2311.15127}.

\bibitem[Bradley \& Nakkiran(2024)Bradley and
  Nakkiran]{bradley2024classifierfreeguidancepredictorcorrector}
Bradley, A. and Nakkiran, P.
\newblock Classifier-free guidance is a predictor-corrector, 2024.
\newblock URL \url{https://arxiv.org/abs/2408.09000}.

\bibitem[Campbell et~al.(2022)Campbell, Benton, Bortoli, Rainforth,
  Deligiannidis, and Doucet]{campbell2022continuoustimeframeworkdiscrete}
Campbell, A., Benton, J., Bortoli, V.~D., Rainforth, T., Deligiannidis, G., and
  Doucet, A.
\newblock A continuous time framework for discrete denoising models, 2022.
\newblock URL \url{https://arxiv.org/abs/2205.14987}.

\bibitem[Campbell et~al.(2024)Campbell, Yim, Barzilay, Rainforth, and
  Jaakkola]{multimodal-flow}
Campbell, A., Yim, J., Barzilay, R., Rainforth, T., and Jaakkola, T.~S.
\newblock Generative flows on discrete state-spaces: Enabling multimodal flows
  with applications to protein co-design.
\newblock In \emph{Forty-first International Conference on Machine Learning,
  {ICML} 2024, Vienna, Austria, July 21-27, 2024}. OpenReview.net, 2024.
\newblock URL \url{https://openreview.net/forum?id=kQwSbv0BR4}.

\bibitem[Carbone et~al.(2023)Carbone, Hua, Coste, and
  Vanden-Eijnden]{carbone2023efficient}
Carbone, D., Hua, M., Coste, S., and Vanden-Eijnden, E.
\newblock Efficient training of energy-based models using jarzynski equality.
\newblock \emph{Advances in Neural Information Processing Systems},
  36:\penalty0 52583--52614, 2023.

\bibitem[Chidambaram et~al.(2024)Chidambaram, Gatmiry, Chen, Lee, and
  Lu]{what-does-guidance-do}
Chidambaram, M., Gatmiry, K., Chen, S., Lee, H., and Lu, J.
\newblock What does guidance do? a fine-grained analysis in a simple setting,
  2024.
\newblock URL \url{https://arxiv.org/abs/2409.13074}.

\bibitem[Clark et~al.(2024)Clark, Vicol, Swersky, and
  Fleet]{clark2024directlyfinetuningdiffusionmodels}
Clark, K., Vicol, P., Swersky, K., and Fleet, D.~J.
\newblock Directly fine-tuning diffusion models on differentiable rewards,
  2024.
\newblock URL \url{https://arxiv.org/abs/2309.17400}.

\bibitem[Cornet et~al.(2024)Cornet, Bartosh, Schmidt, and
  Naesseth]{cornet2024equivariant}
Cornet, F. R.~J., Bartosh, G., Schmidt, M.~N., and Naesseth, C.~A.
\newblock Equivariant neural diffusion for molecule generation.
\newblock In \emph{The Thirty-eighth Annual Conference on Neural Information
  Processing Systems}, 2024.
\newblock URL \url{https://openreview.net/forum?id=40pE5pFhWl}.

\bibitem[Del~Moral \& Penev(2017)Del~Moral and Penev]{Del_Moral_Penev_2017}
Del~Moral, P. and Penev, S.
\newblock \emph{Stochastic processes: From applications to theory}.
\newblock CRC Press, Taylor \& Francis Group, 2017.

\bibitem[Denker et~al.(2024)Denker, Vargas, Padhy, Didi, Mathis, Dutordoir,
  Barbano, Mathieu, Komorowska, and Lio]{DEFT}
Denker, A., Vargas, F., Padhy, S., Didi, K., Mathis, S.~V., Dutordoir, V.,
  Barbano, R., Mathieu, E., Komorowska, U.~J., and Lio, P.
\newblock {DEFT:} efficient finetuning of conditional diffusion models by
  learning the generalised h-transform.
\newblock \emph{CoRR}, abs/2406.01781, 2024.
\newblock \doi{10.48550/ARXIV.2406.01781}.
\newblock URL \url{https://doi.org/10.48550/arXiv.2406.01781}.

\bibitem[Dhariwal \& Nichol(2021)Dhariwal and
  Nichol]{dhariwal2021diffusionmodelsbeatgans}
Dhariwal, P. and Nichol, A.
\newblock Diffusion models beat gans on image synthesis, 2021.
\newblock URL \url{https://arxiv.org/abs/2105.05233}.

\bibitem[Domingo-Enrich et~al.(2025)Domingo-Enrich, Drozdzal, Karrer, and
  Chen]{domingoenrich2025adjointmatchingfinetuningflow}
Domingo-Enrich, C., Drozdzal, M., Karrer, B., and Chen, R. T.~Q.
\newblock Adjoint matching: Fine-tuning flow and diffusion generative models
  with memoryless stochastic optimal control, 2025.
\newblock URL \url{https://arxiv.org/abs/2409.08861}.

\bibitem[Fan et~al.(2023)Fan, Watkins, Du, Liu, Ryu, Boutilier, Abbeel,
  Ghavamzadeh, Lee, and Lee]{NEURIPS2023_fc65fab8}
Fan, Y., Watkins, O., Du, Y., Liu, H., Ryu, M., Boutilier, C., Abbeel, P.,
  Ghavamzadeh, M., Lee, K., and Lee, K.
\newblock Dpok: Reinforcement learning for fine-tuning text-to-image diffusion
  models.
\newblock In Oh, A., Naumann, T., Globerson, A., Saenko, K., Hardt, M., and
  Levine, S. (eds.), \emph{Advances in Neural Information Processing Systems},
  volume~36, pp.\  79858--79885. Curran Associates, Inc., 2023.

\bibitem[Gulrajani \& Hashimoto(2023)Gulrajani and
  Hashimoto]{gulrajani2023likelihoodbaseddiffusionlanguagemodels}
Gulrajani, I. and Hashimoto, T.~B.
\newblock Likelihood-based diffusion language models, 2023.
\newblock URL \url{https://arxiv.org/abs/2305.18619}.

\bibitem[He et~al.(2021)He, Liu, Gao, and
  Chen]{he2021debertadecodingenhancedbertdisentangled}
He, P., Liu, X., Gao, J., and Chen, W.
\newblock Deberta: Decoding-enhanced bert with disentangled attention, 2021.
\newblock URL \url{https://arxiv.org/abs/2006.03654}.

\bibitem[Ho \& Salimans(2022{\natexlab{a}})Ho and
  Salimans]{classifier-free-guidance}
Ho, J. and Salimans, T.
\newblock Classifier-free diffusion guidance, 2022{\natexlab{a}}.
\newblock URL \url{https://arxiv.org/abs/2207.12598}.

\bibitem[Ho \& Salimans(2022{\natexlab{b}})Ho and
  Salimans]{ho2022classifierfreediffusionguidance}
Ho, J. and Salimans, T.
\newblock Classifier-free diffusion guidance, 2022{\natexlab{b}}.
\newblock URL \url{https://arxiv.org/abs/2207.12598}.

\bibitem[Ho et~al.(2020)Ho, Jain, and
  Abbeel]{ho2020denoisingdiffusionprobabilisticmodels}
Ho, J., Jain, A., and Abbeel, P.
\newblock Denoising diffusion probabilistic models, 2020.
\newblock URL \url{https://arxiv.org/abs/2006.11239}.

\bibitem[Hoogeboom et~al.(2022)Hoogeboom, Satorras, Vignac, and
  Welling]{hoogeboom2022equivariantdiffusionmoleculegeneration}
Hoogeboom, E., Satorras, V.~G., Vignac, C., and Welling, M.
\newblock Equivariant diffusion for molecule generation in 3d, 2022.
\newblock URL \url{https://arxiv.org/abs/2203.17003}.

\bibitem[Kerby \& Moon(2024)Kerby and Moon]{training-free-guidance}
Kerby, T.~J. and Moon, K.~R.
\newblock Training-free guidance for discrete diffusion models for molecular
  generation, 2024.
\newblock URL \url{https://arxiv.org/abs/2409.07359}.

\bibitem[Langley(2000)]{langley00}
Langley, P.
\newblock Crafting papers on machine learning.
\newblock In Langley, P. (ed.), \emph{Proceedings of the 17th International
  Conference on Machine Learning (ICML 2000)}, pp.\  1207--1216, Stanford, CA,
  2000. Morgan Kaufmann.

\bibitem[L{\'e}onard(2014)]{leonard2014some}
L{\'e}onard, C.
\newblock Some properties of path measures.
\newblock \emph{S{\'e}minaire de Probabilit{\'e}s XLVI}, pp.\  207--230, 2014.

\bibitem[Li et~al.(2024)Li, Zhao, Wang, Scalia, Eraslan, Nair, Biancalani, Ji,
  Regev, Levine, and Uehara]{li2024derivativefreeguidancecontinuousdiscrete}
Li, X., Zhao, Y., Wang, C., Scalia, G., Eraslan, G., Nair, S., Biancalani, T.,
  Ji, S., Regev, A., Levine, S., and Uehara, M.
\newblock Derivative-free guidance in continuous and discrete diffusion models
  with soft value-based decoding, 2024.
\newblock URL \url{https://arxiv.org/abs/2408.08252}.

\bibitem[Lipman et~al.(2023)Lipman, Chen, Ben-Hamu, Nickel, and
  Le]{lipman2023flowmatchinggenerativemodeling}
Lipman, Y., Chen, R. T.~Q., Ben-Hamu, H., Nickel, M., and Le, M.
\newblock Flow matching for generative modeling, 2023.
\newblock URL \url{https://arxiv.org/abs/2210.02747}.

\bibitem[Lipman et~al.(2024)Lipman, Havasi, Holderrieth, Shaul, Le, Karrer,
  Chen, Lopez-Paz, Ben-Hamu, and Gat]{lipman2024flowmatchingguidecode}
Lipman, Y., Havasi, M., Holderrieth, P., Shaul, N., Le, M., Karrer, B., Chen,
  R. T.~Q., Lopez-Paz, D., Ben-Hamu, H., and Gat, I.
\newblock Flow matching guide and code, 2024.
\newblock URL \url{https://arxiv.org/abs/2412.06264}.

\bibitem[Liu et~al.(2021)Liu, Sap, Lu, Swayamdipta, Bhagavatula, Smith, and
  Choi]{dexperts}
Liu, A., Sap, M., Lu, X., Swayamdipta, S., Bhagavatula, C., Smith, N.~A., and
  Choi, Y.
\newblock {DE}xperts: Decoding-time controlled text generation with experts and
  anti-experts.
\newblock In Zong, C., Xia, F., Li, W., and Navigli, R. (eds.),
  \emph{Proceedings of the 59th Annual Meeting of the Association for
  Computational Linguistics and the 11th International Joint Conference on
  Natural Language Processing (Volume 1: Long Papers)}, pp.\  6691--6706,
  Online, August 2021. Association for Computational Linguistics.
\newblock \doi{10.18653/v1/2021.acl-long.522}.
\newblock URL \url{https://aclanthology.org/2021.acl-long.522/}.

\bibitem[Lou et~al.(2024)Lou, Meng, and Ermon]{sedd}
Lou, A., Meng, C., and Ermon, S.
\newblock Discrete diffusion language modeling by estimating the ratios of the
  data distribution, 2024.
\newblock URL \url{https://openreview.net/forum?id=71mqtQdKB9}.

\bibitem[Martino et~al.(2016)Martino, Elvira, and Louzada]{partal-resampling}
Martino, L., Elvira, V., and Louzada, F.
\newblock Weighting a resampled particle in sequential monte carlo.
\newblock In \emph{2016 IEEE Statistical Signal Processing Workshop (SSP)},
  pp.\  1--5, 2016.
\newblock \doi{10.1109/SSP.2016.7551711}.

\bibitem[Nisonoff et~al.(2024)Nisonoff, Xiong, Allenspach, and
  Listgarten]{unlock_guidance}
Nisonoff, H., Xiong, J., Allenspach, S., and Listgarten, J.
\newblock Unlocking guidance for discrete state-space diffusion and flow
  models, 2024.
\newblock URL \url{https://arxiv.org/abs/2406.01572}.

\bibitem[Podell et~al.(2023)Podell, English, Lacey, Blattmann, Dockhorn,
  Müller, Penna, and Rombach]{podell2023sdxlimprovinglatentdiffusion}
Podell, D., English, Z., Lacey, K., Blattmann, A., Dockhorn, T., Müller, J.,
  Penna, J., and Rombach, R.
\newblock Sdxl: Improving latent diffusion models for high-resolution image
  synthesis, 2023.
\newblock URL \url{https://arxiv.org/abs/2307.01952}.

\bibitem[Rombach et~al.(2022)Rombach, Blattmann, Lorenz, Esser, and
  Ommer]{rombach2022highresolutionimagesynthesislatent}
Rombach, R., Blattmann, A., Lorenz, D., Esser, P., and Ommer, B.
\newblock High-resolution image synthesis with latent diffusion models, 2022.
\newblock URL \url{https://arxiv.org/abs/2112.10752}.

\bibitem[Ronneberger et~al.(2015)Ronneberger, Fischer, and
  Brox]{ronneberger2015unetconvolutionalnetworksbiomedical}
Ronneberger, O., Fischer, P., and Brox, T.
\newblock U-net: Convolutional networks for biomedical image segmentation,
  2015.
\newblock URL \url{https://arxiv.org/abs/1505.04597}.

\bibitem[Shi et~al.(2024)Shi, Han, Wang, Doucet, and Titsias]{diffusion-jiaxin}
Shi, J., Han, K., Wang, Z., Doucet, A., and Titsias, M.
\newblock Simplified and generalized masked diffusion for discrete data.
\newblock In \emph{The Thirty-eighth Annual Conference on Neural Information
  Processing Systems}, 2024.
\newblock URL \url{https://openreview.net/forum?id=xcqSOfHt4g}.

\bibitem[Song \& Ermon(2020)Song and
  Ermon]{song2020generativemodelingestimatinggradients}
Song, Y. and Ermon, S.
\newblock Generative modeling by estimating gradients of the data distribution,
  2020.
\newblock URL \url{https://arxiv.org/abs/1907.05600}.

\bibitem[Song et~al.(2021)Song, Sohl-Dickstein, Kingma, Kumar, Ermon, and
  Poole]{song2021scorebasedgenerativemodelingstochastic}
Song, Y., Sohl-Dickstein, J., Kingma, D.~P., Kumar, A., Ermon, S., and Poole,
  B.
\newblock Score-based generative modeling through stochastic differential
  equations, 2021.
\newblock URL \url{https://arxiv.org/abs/2011.13456}.

\bibitem[Uehara et~al.(2025)Uehara, Zhao, Wang, Li, Regev, Levine, and
  Biancalani]{uehara2025inferencetimealignmentdiffusionmodels}
Uehara, M., Zhao, Y., Wang, C., Li, X., Regev, A., Levine, S., and Biancalani,
  T.
\newblock Inference-time alignment in diffusion models with reward-guided
  generation: Tutorial and review, 2025.
\newblock URL \url{https://arxiv.org/abs/2501.09685}.

\bibitem[Vargas et~al.(2023)Vargas, Grathwohl, and Doucet]{vargas2023denoising}
Vargas, F., Grathwohl, W., and Doucet, A.
\newblock Denoising diffusion samplers.
\newblock \emph{arXiv preprint arXiv:2302.13834}, 2023.

\bibitem[Vargas et~al.(2024)Vargas, Nusken, Padhy, and
  Blessing]{nusken2024transport}
Vargas, F., Nusken, N., Padhy, S., and Blessing, D.
\newblock Transport meets variational inference: Controlled monte carlo
  diffusions.
\newblock In \emph{The Twelfth International Conference on Learning
  Representations: ICLR 2024}, 2024.

\bibitem[Venkatraman et~al.(2024)Venkatraman, Jain, Scimeca, Kim, Sendera,
  Hasan, Rowe, Mittal, Lemos, Bengio, Adam, Rector-Brooks, Bengio, Berseth, and
  Malkin]{venkatraman2024amortizing}
Venkatraman, S., Jain, M., Scimeca, L., Kim, M., Sendera, M., Hasan, M., Rowe,
  L., Mittal, S., Lemos, P., Bengio, E., Adam, A., Rector-Brooks, J., Bengio,
  Y., Berseth, G., and Malkin, N.
\newblock Amortizing intractable inference in diffusion models for vision,
  language, and control.
\newblock In \emph{The Thirty-eighth Annual Conference on Neural Information
  Processing Systems}, 2024.
\newblock URL \url{https://openreview.net/forum?id=gVTkMsaaGI}.

\bibitem[Vignac et~al.(2023)Vignac, Krawczuk, Siraudin, Wang, Cevher, and
  Frossard]{vignac2023digress}
Vignac, C., Krawczuk, I., Siraudin, A., Wang, B., Cevher, V., and Frossard, P.
\newblock Digress: Discrete denoising diffusion for graph generation.
\newblock In \emph{The Eleventh International Conference on Learning
  Representations}, 2023.
\newblock URL \url{https://openreview.net/forum?id=UaAD-Nu86WX}.

\bibitem[Watson et~al.(2023)Watson, Juergens, Bennett, Trippe, Yim, Eisenach,
  Ahern, Borst, Ragotte, Milles, et~al.]{watson2023novo}
Watson, J.~L., Juergens, D., Bennett, N.~R., Trippe, B.~L., Yim, J., Eisenach,
  H.~E., Ahern, W., Borst, A.~J., Ragotte, R.~J., Milles, L.~F., et~al.
\newblock De novo design of protein structure and function with rfdiffusion.
\newblock \emph{Nature}, 620\penalty0 (7976):\penalty0 1089--1100, 2023.

\bibitem[Wu et~al.(2024)Wu, Trippe, Naesseth, Blei, and
  Cunningham]{wu2024practical}
Wu, L., Trippe, B., Naesseth, C., Blei, D., and Cunningham, J.~P.
\newblock Practical and asymptotically exact conditional sampling in diffusion
  models.
\newblock \emph{Advances in Neural Information Processing Systems}, 36, 2024.

\bibitem[Wu et~al.(2016)Wu, Schuster, Chen, Le, Norouzi, Macherey, Krikun, Cao,
  Gao, Macherey, Klingner, Shah, Johnson, Liu, Łukasz Kaiser, Gouws, Kato,
  Kudo, Kazawa, Stevens, Kurian, Patil, Wang, Young, Smith, Riesa, Rudnick,
  Vinyals, Corrado, Hughes, and Dean]{wu2016googlesneuralmachinetranslation}
Wu, Y., Schuster, M., Chen, Z., Le, Q.~V., Norouzi, M., Macherey, W., Krikun,
  M., Cao, Y., Gao, Q., Macherey, K., Klingner, J., Shah, A., Johnson, M., Liu,
  X., Łukasz Kaiser, Gouws, S., Kato, Y., Kudo, T., Kazawa, H., Stevens, K.,
  Kurian, G., Patil, N., Wang, W., Young, C., Smith, J., Riesa, J., Rudnick,
  A., Vinyals, O., Corrado, G., Hughes, M., and Dean, J.
\newblock Google's neural machine translation system: Bridging the gap between
  human and machine translation, 2016.
\newblock URL \url{https://arxiv.org/abs/1609.08144}.

\bibitem[Zhang* et~al.(2020)Zhang*, Kishore*, Wu*, Weinberger, and
  Artzi]{bert-score}
Zhang*, T., Kishore*, V., Wu*, F., Weinberger, K.~Q., and Artzi, Y.
\newblock Bertscore: Evaluating text generation with bert.
\newblock In \emph{International Conference on Learning Representations}, 2020.
\newblock URL \url{https://openreview.net/forum?id=SkeHuCVFDr}.

\end{thebibliography}
\bibliographystyle{icml2025}

%%%%%%%%%%%%%%%%%%%%%%%%%%%%%%%%%%%%%%%%%%%%%%%%%%%%%%%%%%%%%%%%%%%%%%%%%%%%%%%
%%%%%%%%%%%%%%%%%%%%%%%%%%%%%%%%%%%%%%%%%%%%%%%%%%%%%%%%%%%%%%%%%%%%%%%%%%%%%%%
% APPENDIX
%%%%%%%%%%%%%%%%%%%%%%%%%%%%%%%%%%%%%%%%%%%%%%%%%%%%%%%%%%%%%%%%%%%%%%%%%%%%%%%
%%%%%%%%%%%%%%%%%%%%%%%%%%%%%%%%%%%%%%%%%%%%%%%%%%%%%%%%%%%%%%%%%%%%%%%%%%%%%%%
\newpage
\appendix
\onecolumn

\section{Related work}
Diffusion models have emerged as a powerful class of generative models that can be interpreted through various theoretical frameworks, including score-based modeling \citep{song2020generativemodelingestimatinggradients, song2021scorebasedgenerativemodelingstochastic}, flow matching \citep{lipman2023flowmatchinggenerativemodeling, lipman2024flowmatchingguidecode}, variational methods \citep{ho2020denoisingdiffusionprobabilisticmodels} and stochastic interpolant \citep{albergo2023buildingnormalizingflowsstochastic, albergo2023stochasticinterpolantsunifyingframework}. Their versatility has led to successful applications across diverse domains, from image and video generation \citep{rombach2022highresolutionimagesynthesislatent, podell2023sdxlimprovinglatentdiffusion, blattmann2023stablevideodiffusionscaling}  to molecular design \citep{cornet2024equivariant, vignac2023digress, hoogeboom2022equivariantdiffusionmoleculegeneration}, protein design \citep{watson2023novo}, and text generation \citep{diffusion-jiaxin, sedd, gulrajani2023likelihoodbaseddiffusionlanguagemodels}, where controlled generation is often crucial. \\
Two primary approaches have been developed for controlled generation: classifier guidance \citep{dhariwal2021diffusionmodelsbeatgans} and classifier-free guidance \citep{ho2022classifierfreediffusionguidance}. Classifier guidance incorporates gradients from a trained classifier to steer the generation process, while CFG eliminates the need for a separate classifier by jointly training conditional and unconditional models. For scenarios with limited data where training classifiers or conditional models is impractical, DEFT \citep{DEFT} demonstrates that fine-tuning a small network can learn classifier scores directly, enabling guidance of unconditional models.
However, recent theoretical analyses \citep{what-does-guidance-do, bradley2024classifierfreeguidancepredictorcorrector} have revealed fundamental limitations of guidance methods. Specifically, as the guidance temperature parameter increases to strengthen conditioning, these methods fail to sample from their intended tilted target distributions. This limitation is particularly relevant for applications requiring precise control over generated samples.\\
In the discrete setting, several approaches have been proposed to adapt guidance methods. \citet{unlock_guidance} adapt Classifier Free Guidance, \citet{vignac2023digress} use Taylor expansions to approximate guidance terms with a property-predicting regressor. \citet{li2024derivativefreeguidancecontinuousdiscrete} introduce SVDD, integrating value functions for reward-based sampling without fine-tuning. \citet{training-free-guidance} develop a training-free guidance framework for discrete data generation. \\
Recent works propose finetuning approaches using reinforcement learning to sample from tempered distributions \citep{domingoenrich2025adjointmatchingfinetuningflow, venkatraman2024amortizing, NEURIPS2023_fc65fab8, black2024trainingdiffusionmodelsreinforcement, clark2024directlyfinetuningdiffusionmodels, uehara2025inferencetimealignmentdiffusionmodels}. \citet{black2024trainingdiffusionmodelsreinforcement} introduce Denoising Diffusion Policy Optimization (DDPO), which reformulates the denoising process as a multi-step decision problem. \citet{clark2024directlyfinetuningdiffusionmodels} propose Direct Reward Fine-tuning (DRaFT), demonstrating that backpropagation through the entire sampling procedure can effectively optimize differentiable reward functions. \citet{NEURIPS2023_fc65fab8}  develop DPOK, which combines policy optimization with KL regularization for fine-tuning text-to-image models, showing improvements in both image-text alignment and image quality compared to supervised fine-tuning approaches.

\section{A Primer on Continuous-Time Markov Chains} \label{app:primer-ctmc}
\subsection{Definition}
A Continuous-Time Markov Chain $\{X_t\}_{t\in[0,1]}$ on finite state space $\mathcal{X} = \{1, \dots, |\mathcal{X}|\}$ is collection of time-indexed random variables taking values on $\mathcal{X}$. A CTMC $\{X_t\}_{t\in[0,1]}$ obeys the Markov property, that is, for all $A \subseteq \mathcal{X}$, $n\in \mathbb{N}$, $t_1, \dots, t_n \in [0, 1]$, and $0 \leq x_1 < \dots < x_n \leq 1 \in \mathcal{X}$:
\begin{align}
    \P(X_t \in A | X_{t_1} = x_1, \dots X_{t_n} = x_n) = \P(X_t \in A | X_{t_n} = x_n).
\end{align}
The initial distribution $p_0(x_0)$ and transition probability $p_{t|s}(x_t|x_s)$ are defined as,
\begin{align}
p_0(x_0) = \P(X_0 = x_0)  &&  p_{t|s}(x_t|x_s) = \P(X_t = x_t, X_s = x_s).
\end{align}
Notably, a CTMC is \textbf{càdlàg} by convention, that is for every $t \geq 0$:
\begin{enumerate}
    \item $\lim_{s \to t^+} X(s) = X(t)$  (right-continuous at $t$),
    \item $\lim_{s \to t^-} X(s)$ exists (finite left-hand limit at $t$).
\end{enumerate}
\subsection{The Transition Rate Matrix}
One way to characterise a CTMC is through the time-dependent \textbf{transition rate matrix} $R_t$ (also known as the \textbf{generator}), defined as,
\begin{align}
    \forall x, y\in\mathcal{X}. \;R_t(x, y) = \lim_{\deltat\to 0}\frac{p_{t+\deltat|t}(x_{t+\deltat}|x_t) - \delta_{x,y}}{\deltat}
\end{align}
where $\delta_{a,b}$ is 1 if $a = b$ and $0$ if otherwise.

By definition, a first-order approximation of the transition probability is:
\begin{align}
    p_{t+\deltat|t}(x_{t+\deltat|t}|x_t) = \delta_{x_{t+\deltat,x_t}} + R_t(x_t, x_{t+\deltat}) \deltat + o(\deltat).
\end{align}
By noticing that $\sum_{y\in\mathcal{X}} p_{t+\deltat|t}(y | x) = 1$, we derive the mass conservation property of $R_t$,
\begin{align}
    R_t(x, x) = -\sum_{y\in\mathcal{X}: x\neq x} R_t(x, y)
\end{align}
By convention, we define $R_t(x)$ as,
\begin{align}
    R_t(x) = \sum_{y\in\mathcal{X}: x\neq x} R_t(x, y).
\end{align}
\subsection{The Kolmogorov Equations}
In this section, we outline both Kolmogorov Forward and Backward Equation that govern respectively the change of the probability mass, and expectations of some function over time.

\textbf{Kolmogorov-Forward Equation.} \; The Kolmogorov Forward Equation (or the continuity equation) governs the change of probability mass over time. It states that the probability mass $p_t$ satisfies the following equation:
\begin{align}
    \forall 0 \leq t \leq 1. \; \partial_t p_t(x_t) = \underbrace{\sum_{y\in\mathcal{X}:x_t\neq y} R_t(y, x_t) p_t(y)}_\text{incoming probability mass} - \underbrace{\sum_{y\in\mathcal{X}:x_t\neq y} R_t(x_t, y) p_t(x_t)}_\text{outgoing probability mass}
\end{align}
In other words, the rate of change of probability mass equals the difference between incoming and outgoing probability flows. In matrix notation, this equation is:
\begin{align}
    \partial_t p_t = R_t^\intercal p_t
\end{align}

\textbf{Kolmogorov Backward Equation} \; Given a function $h : \mathcal{X} \to \R$, the Kolmogorov Backward Equation states that the function $u_t(x)$ defined as,
\begin{align}
    u_t(x) = \E[h(X_1) | X_t = x],
\end{align}
satisfies the ordinary differential equation,
\begin{align}
    \partial_t u_t(x) = -\sum_{y} R_t(x, y) u_t(y),
\end{align}
writing $u_t$ as a column vector, this can be written in matrix form,
\begin{align}
    \partial_t u_t = -R_t^\intercal u_t.
\end{align}

\subsection{Reverse-time CTMC}
Time-reversed CTMC is a core concept of diffusion model. A reverse-time CTMC $\{Y_t\}_{\in[0,1]}$ with time-dependent rate $R_t$ is defined as $Y_t = \hat{Y}_{1-t}$ where $\{\hat{Y_t}\}_{t\in[0,1]}$ is a forward-time CTMC with transition rate $R_{1-t}$. The reverse-time Kolmogorov Equations are then obtained by applying a change of variable $t\mapsto 1-t$.

\section{Proof for \Cref{sec:cond-guide}}
In this section, we prove \Cref{prop:true-tempered}.

\textbf{\Cref{prop:true-tempered}.}\; Let $M_t[q]$ denote the evolution of probability mass $q$ up to time $t$ under the corrupting process. Define $p_t^{\alpha,\text{true}}$ as:
\begin{align*}
p_t^{\alpha,\text{true}} = M_t\big[p_0(\cdot) \, p_0(\zeta | \cdot)^\alpha / \gZ_\alpha\big].
\end{align*}
Then,
\begin{align*}
p_t^{\alpha,\text{true}}(x_t) \propto p_t(x_t) \mathbb{E}\big[p_0(\zeta | X_0)^\alpha | X_t = x_t\big].
\end{align*}
The true tempered rate matrix $R_t^{\alpha,\text{true}}$ given by:
\begin{align*}
\forall x \neq y, \; R_t^{\alpha,\text{true}}(x, y | \zeta) &= R_t(x, y) \tfrac{\mathbb{E}[p_0(\zeta | X_0)^\alpha | X_t = y]}{\mathbb{E}[p_0(\zeta | X_0)^\alpha | X_t = x]}, \\ 
\forall x, \; R_t^{\alpha,\text{true}}(x, x | \zeta) &= -\sum_{y \neq x} R_t^{\alpha,\text{true}}(x, y | \zeta),
\end{align*}
is time-reversal of the corruption process that stats from $p_0(\cdot) p_0(\zeta|\cdot)^\alpha / \gZ_\alpha$
\begin{proof}
Recall that $\{X_t\}_{t\in[0,1]}$ is a reversed-time CTMC corresponding to the unconditonal discrete diffusion model. Let $\{p_{t|s}^\text{corrupt}(x_t|x_s)\}_{t>s}$ and $\{p_{t|s}(x_t|x_s)\}_{t<s}$ be the transition probability of the forward-time corrupting process and reverse-time generative process. We first show $p_t^\alpha(x_t) \propto p_t(x_t) \mathbb{E}\big[p_0(\zeta | X_0)^\alpha | X_t = x_t\big]$. By definition,
\begin{align}
    p_t^{\alpha,\text{true}}(x_t) &= M_t[p_0(\cdot)p(\zeta|\cdot)^\alpha / \gZ_\alpha] \\ 
                    &\propto \sum_{x_0} p_{t|0}^\text{corrupt}(x_t|x_0) p_0(x_0)p(\zeta|x_0)^\alpha \\ 
                    &= \sum_{x_0} p_{0|t}(x_0|x_t) \frac{p_t(x_t)}{p_0(x_0)}p_0(x_0)p(\zeta|x_0)^\alpha \\ 
                    &= p_t(x_t) \sum_{x_0}p_{0|t}(x_0|x_t) p(\zeta|x_0)^\alpha\\
                    &= p_t(x_t) \E[p(\zeta|X_0)^\alpha|X_t=x_t].
\end{align}

Then it follow from standard results of time-reversal of the CTMC such as that in \cite{sedd}, the rate matrix of the time-reversal can be written as:
\begin{align}
\forall x\neq y. \quad    R_t^{\alpha,\text{true}}(x, y | \zeta) = R_t(x, y) \frac{p_t^{\alpha,\text{true}}(y)}{p_t^{\alpha,\text{true}}(x)} =  R_t^{\alpha,\text{true}}(x, y | \zeta) &= R_t(x, y) \tfrac{\mathbb{E}[p_0(\zeta | X_0)^\alpha | X_t = y]}{\mathbb{E}[p_0(\zeta | X_0)^\alpha | X_t = x]}
\end{align}
\end{proof}
\section{Proof of \Cref{prop:cont-resampling}} \label{sec:proof-cont-resampling}

\textbf{\Cref{prop:cont-resampling}.} \; 
Let proposal $\{Y_t\}_{t\in[0,1]}$ be reverse-time CTMC  with rate matrix $Q_t$ and initial distribution $p_1(\cdot)$. Define $\{W_t\}_{t\in[0,1]}$ by 
\begin{align*}
W_t = \frac{\mathrm{d} \mathbb{P}^{\mathrm{base}}_t}{\mathrm{d}  \mathbb{Q}_t} \left(\frac{\mathrm{d} \mathbb{P}^{ \alpha=1}_t }{\mathrm{d}\mathbb{P}^{\mathrm{base}}_t } \right)^\alpha, %\!\!\!\!(Y_t),
\end{align*}
where we have $\sP_t^{\text{base}}=\mathrm{Law}{\{X_\tau\}_{\tau\in[t,1]}}$, $\sP^{\alpha=1}=\mathrm{Law}{\{X_\tau\mid\zeta\}_{\tau\in[t,1]}}$,  $\sQ = \mathrm{Law}{\{Y_\tau\}_{\tau\in[t,1]}}$.
Then,
\begin{align*}
\ln \frac{\mathrm{d} \mathbb{P}_t}{\mathrm{d} \mathbb{Q}_t}^{\!\mathrm{ base}} 
 =  &\displaystyle\sum_{\substack{ \tau  \geq t\\ Y_{\tau^+}\neq Y_\tau}} 
\!\!\ln R_t(Y_{\tau^+}, Y_\tau) - \ln Q_t(Y_{\tau^+}, Ys_\tau) 
+ \int_{1}^t Q_\tau(Y_t) - R_\tau(Y_t) \, \mathrm{d} \tau
\end{align*}
and
\begin{align*}
\ln\! \frac{\mathrm{d} \mathbb{P}_t^{ \alpha=1} }{\mathrm{d}\mathbb{P}_t^{\mathrm{base}} } 
\!= &\!\!\!\!\!\displaystyle\sum_{\substack{ \tau  \geq t\\ Y_{\tau^+}\neq Y_\tau}} 
\!\!\!\!\ln R_t^{\alpha=1}\!(Y_{\tau^+}, Y_\tau|\zeta) \!- \!\ln R_t(Y_{\tau^+}, Y_\tau) 
+\!\!\int_{1}^t  \!\!\! R_\tau(Y_\tau) - R^{\alpha=1}_\tau(Y_\tau|\zeta) \mathrm{d} \tau.
\end{align*}
For any $p_t^{\alpha}$-integrable function $h$ and time $t \in [0, 1]$,
\begin{align*}
    \E_{x \sim p_t^\alpha}[h(x)] = \frac{\E[W_t \cdot h(Y_t)]}{\E[W_t]}.
\end{align*}

\begin{proof}
Via the Radon-Nikodym Theorem, we seek to compute the importance weight (i.e. the Radon-Nikodym derivative \footnote{We consider all RND's evaluated at $Y_t$ that is when we write $\frac{\dd \P_t'}{\dd \Q'_t}$ it is short for $\frac{\dd \P_t'}{\dd \Q'_t}(Y_t)$.})
\begin{align}
\frac{\mathrm{d} \mathbb{P}^{ \alpha}_t}{\dd \sQ_t} = \frac{\mathrm{d} \mathbb{P}^{\!\mathrm{ base}}_t}{\mathrm{d} \mathbb{Q}_t} \cdot \frac{\mathrm{d} \mathbb{P}^{ \alpha}_t}{\mathrm{d}\mathbb{P}_t^{\mathrm{base}} } =\frac{\mathrm{d} \mathbb{P}^{\!\mathrm{ base}}_t}{\mathrm{d} \mathbb{Q}_t} \cdot p_t(\zeta|Y_t)^\alpha, 
\end{align}
where $\frac{\mathrm{d} \mathbb{P}^{ \alpha}_t}{\mathrm{d}\mathbb{P}_t^{\mathrm{base}} } = p_t(\zeta|Y_t)^\alpha$ follows by construction of $\mathbb{P}^{ \alpha}_t$ and a direct application of the disintegration theorem \citep{leonard2014some} (i.e. the product rule),
% Notice that by the Radon-Nikodym Theorem, that one possible importance weight $W_t$ is:
% \begin{align}
% \frac{\mathrm{d} \mathbb{P}^{\!\mathrm{ base}}_t}{\mathrm{d} \mathbb{Q}_t} \cdot p_t(\zeta|Y_t)^\alpha.
% \end{align}
then considering $\alpha=1$ it follows that $p_t(\zeta|Y_t)^\alpha=\left(\frac{\mathrm{d} \mathbb{P}^{\alpha=1}_t}{\mathrm{d}\mathbb{P}_t^{\mathrm{base}}}\right)^\alpha$, and thus the importance weights between our proposal and the tilted path measure reduce to:
\begin{align}
W_t = \frac{\mathrm{d} \mathbb{P}^{ \alpha}_t}{\dd \sQ_t} = \frac{\mathrm{d} \mathbb{P}^{\mathrm{base}}_t}{\mathrm{d}  \mathbb{Q}_t} \left(\frac{\mathrm{d} \mathbb{P}^{ \alpha=1}_t }{\mathrm{d}\mathbb{P}^{\mathrm{base}_t} } \right)^\alpha, \label{eq:weightchainrule} %\!\!\!\!(Y_t), 
\end{align}
Finally, the Radon-Nikodym derivative between two arbitrary reverse-time CTMC path measures $\P_t'$ and $\Q_t'$ with transition rate matrix $R'_\tau$ and $Q'_\tau$ evaluated at the path $Y$, is known to be the following Appendix C.1 of \citep{multimodal-flow}, that is,
\begin{align}\label{eq:ctmcrnd}
     \frac{\dd \P_t'}{\dd \Q'_t} = \frac{\exp(-\int_{1}^t R'_\tau(Y_t)  \, \mathrm{d} \tau)}{\exp(-\int_{1}^t Q'_\tau(Y_t)  \, \mathrm{d} \tau)}\prod_{\substack{ \tau  \geq t\\ Y_{\tau^+}\neq Y_\tau}} 
\!\!\frac{ R_t'(Y_{\tau^+}, Y_\tau)}{ Q_t'(Y_{\tau^+}, Y_\tau)}
\end{align}
We refer the readers to \citep{multimodal-flow} for an accessible sketch of this result, which follows \citep{Del_Moral_Penev_2017}. 
Then, substituting Equation \ref{eq:ctmcrnd}  into the RNDs of Equation \ref{eq:weightchainrule} concludes the proof.
\end{proof}

\subsection{It\^{o} Process Counterpart} \label{sec:sde_prop}

The core result in Proposition \Cref{prop:cont-resampling} is in noticing that $p_t(\zeta|Y_t)^\alpha=\left(\frac{\mathrm{d} \mathbb{P}^{\alpha=1}_t}{\mathrm{d}\mathbb{P}_t^{\mathrm{base}}}\right)^\alpha$ and decomposing the IS weight via the chain rule and applying Girsanovs Theorem. 

Due to the very modular nature of our result our proposition seamlessly extends to Stochastic Differential Equations.
\begin{proposition} \label{prop:sdecounterprop}
    Let proposal $\{Y_t\}_{t\in[0,1]}$ be a reverse-time SDE with drift $b_t(x)$, diffusion coeficient $g_t$ and initial distribution $p_1(\cdot)$ and $X_t$ also be a reverse-time SDE with drift $f_t$, diffusion coeficient $g_t$ and initial distribution $p_1(\cdot)$. Define $\{W_t\}_{t\in[0,1]}$ by 
\begin{align} \label{eq:chaintmp}
W_t = \frac{\mathrm{d} \mathbb{P}^{\mathrm{base}}_t}{\mathrm{d}  \mathbb{Q}_t} \left(\frac{\mathrm{d} \mathbb{P}^{ \alpha=1}_t }{\mathrm{d}\mathbb{P}^{\mathrm{base}}_t } \right)^\alpha, %\!\!\!\!(Y_t),
\end{align}
where we have $\sP_t^{\text{base}}=\mathrm{Law}{\{X_\tau\}_{\tau\in[t,1]}}$, $\sP^{\alpha=1}=\mathrm{Law}{\{X_\tau \mid\zeta\}_{\tau\in[t,1]}}$,  $\sQ = \mathrm{Law}{\{Y_\tau^{b,p_1}\}_{\tau\in[t,1]}}$.
Then,
\begin{align*}
\ln \frac{\mathrm{d} \mathbb{P}_t}{\mathrm{d} \mathbb{Q}_t}^{\!\mathrm{ base}} 
 = \int_t^1 \frac{1}{g_\tau^2}(f_\tau(Y_\tau)- b_\tau(Y_\tau))^\top \overleftarrow{\dd} Y_\tau + \int_t^1 \frac{1}{2g_\tau^2}(||^2f_\tau(Y_\tau)||^2 -||b_\tau(Y_\tau)||^2) \dd\tau
\end{align*}
and
\begin{align*}
\ln\! \frac{\mathrm{d} \mathbb{P}_t^{ \alpha=1} }{\mathrm{d}\mathbb{P}_t^{\mathrm{base}} } 
\!=  \int_t^1 \frac{1}{g_\tau^2}( f_\tau^{\cdot|\zeta}(Y_\tau)-f_\tau(Y_\tau))^\top \overleftarrow{\dd} Y_\tau + \int_t^1 \frac{1}{2g_\tau^2}(||f^{\cdot|\zeta}_\tau(Y_\tau)||^2 -||f_\tau(Y_\tau)||^2) \dd\tau
\end{align*}
Where $f^{\cdot|\zeta}_\tau(y_\tau) = f_\tau(y_\tau) - g_\tau^2 \nabla \ln p_\tau(\zeta| y_\tau)$ as per  \citep[Proposition 2.2]{DEFT}.

For any $p_t^{\alpha}$-integrable function $h$ and time $t \in [0, 1]$,
\begin{align*}
    \E_{x \sim p_t^\alpha}[h(x)] = \frac{\E[W_t \cdot h(Y_t)]}{\E[W_t]}.
\end{align*}
\end{proposition}
\begin{proof}
Equation \ref{eq:chaintmp} follows directly from \Cref{prop:cont-resampling}. Then, for the RNDs between two reverse time SDEs, we use Equation 64 in \cite{nusken2024transport}. Note for the VP-SDE case; these expressions can be simplified further as done in \citep[Equation 9]{vargas2023denoising}.  Note as with the rest of our results, we require that $p_1(y_1|\zeta) = p_1(y_1)$ which is approximately true in the limit for VP-SDE as $p_1(\cdot|\zeta) \approx \mathcal{N}(0,I)$.
\end{proof}

\subsection{Alternative Proof of \Cref{prop:cont-resampling} without Radon-Nikodym Theorem} \label{sec:alternative-proof}
In this appendix, we present an equivalent statement of \Cref{prop:cont-resampling} and a proof that does not rely on the Radon-Nikodym Theorem.

\textbf{\Cref{prop:cont-resampling}.} \; Let proposal $\{Y_t\}_{t\in[0,1]}$ be a reverse-time CTMC with rate matrix $Q_t$ and initial distribution $p_1(\cdot)$. Define $\{W_t\}_{t\in[0,1]}$ by,
\begin{align*}
W_t &= \exp(A_t), \notag \\ 
A_0 &= 0, \notag \\
A_t &= \sum_{\substack{t \leq \tau \\ Y_{\tau^+}\neq Y_\tau}} \left[\ln R_\tau(Y_{\tau^+}, Y_\tau) - \ln Q_\tau(Y_{\tau^+}, Y_\tau)\right]
     + \sum_{\substack{t \leq \tau \\ Y_{\tau^+}\neq Y_\tau}} \alpha \ln (R^{\alpha=1}_\tau(Y_{\tau^+}, Y_\tau|\zeta)/R_\tau(Y_{\tau^+}, Y_\tau)) \\
     &+ \int_{1}^t \left[Q_\tau(Y_t) - R_\tau(Y_t) +  \alpha R_\tau(Y_\tau|\zeta) - \alpha R^{\alpha=1}_\tau(Y_\tau|\zeta) \right] \, \dd \tau.
\end{align*}
For any $p_t$-integrable function $h$ and time $t \in [0, 1]$,
\begin{align*}
    \E_{x \sim p_t^\alpha}[h(x)] = \frac{\E[W_t \cdot h(Y_t)]}{\E[W_t]}.
\end{align*}
where the expectation is taken over the law of $(Y_t, W_t)$.
\begin{proof}
We closely follow the sketches of \cite{carbone2023efficient,albergo2024nets}. Assume $p_1(x_1) = p_1(x_1|\zeta)$. We first show that the coupled system $(Y_t, A_t)$ with $A_t$ defined as,
\begin{align}
W_0 &= 0 \\  
A_t &= \sum_{\substack{t \leq \tau \\ Y_{\tau^+}\neq Y_\tau}} \left[\ln R_t(Y_{\tau^+}, Y_\tau) - \ln Q_t(Y_{\tau^+}, Y_\tau)\right]+ \sum_{\substack{t \leq \tau \\ Y_{\tau^+}\neq Y_\tau}} \alpha \left[\ln p_t(\zeta|Y_\tau) - \ln p_t(\zeta|Y_{\tau^+})\right] \\
&+ \int_{1}^t \left[R_\tau(Y_t) - Q_\tau(Y_t) - \alpha \partial_t \ln p_t(\zeta | Y_t)\right] \, \dd \tau,
\end{align}
satisfies,
\begin{align}
\E_{x\sim p^\alpha_t}[h(x)] = \frac{\E[e^{A_t} h(Y_t)]}{\E[e^{A_t}]},
\end{align}
Write $\Delta A_t(x, y) = \ln R_t(x, y) - \ln Q_t(x, y) + \alpha(\ln p_t(\zeta |y) - \ln p_t(\zeta | x))$. Define $f_t(y, a)$ to be the joint density of $(Y_t, A_t)$, the joint density is governed by the (reverse-time) continuity equation,
\begin{align}
\partial_t f_t(y, a) &= -\left[\sum_{z \neq y} Q_t(z, y) f_t(z, a - \Delta A(z, y)) - f_t(y, a) Q_t(y) +  \left[ Q_t(y) - R_t(y) - \alpha \partial_t \ln p_t(\zeta | y)\right]\partial_a f_t(y, a)\right],  \nonumber\\  
f_1(y, a) &= \delta{(a)} p_1(x_1)  \label{eq:extkol}
\end{align}
\end{proof}
Define $g_t(y) = \int_{\R}e^a f_t(y, a) da$. Then, via dominated convergence theorem (i.e. $\partial_t g_t(y) = \int_{\R} e^a \partial_tf_t(y, a) da$) and substituting $\partial_tf_t(y, a)$ from Equation \ref{eq:extkol}, it follows that
% Notice that
% \begin{align}
%  \partial_t g_t(y) = \int_{\R} e^a \partial_tf_t(y, a) da.
% \end{align}
% Then notice,
\begin{align} 
\partial_t g_t(y) = \int_{\R} e^a\sum_{z \neq y} Q_t(z, y) f_t(z, a - \Delta A(z, y)) da
\end{align}
By a change of variable $a' = a - \Delta A(z, y)$, then,
\begin{align}
\int_{\R} e^a\sum_{z \neq y} Q_t(z, y) f_t(z, a - \Delta A(z, y)) da
 &= \int_{\R} e^{a'} \sum_{z \neq y} e^{\Delta A(z, y)} Q_t(z, y) f_t(z, a') da'  \\
 &= \sum_{z\neq y} R_t(z, y) \cdot \frac{p_t(\zeta|y)^\alpha}{p_t(\zeta|z)^\alpha} \cdot g_t(z)
\end{align}
And finally, by integration by parts, and noting that $\int_\R e^a f_t(y, a) da = Q_t(y) g_t(y)$, we have
\begin{align}
\int_\R e^a \left[Q_t(y) - R_t(y) + \alpha \partial_t \ln p_t(\zeta | y)\right]\partial_a f_t(y, a) da = \left[Q_t(y) - R_t(y) - \alpha \partial_t \ln p_t(\zeta | y)\right] g_t(y).
\end{align}
Combining the three cases, we have:
\begin{align}
g_1(y) &= p_1(y) \\ 
\partial_t g_t(y) &= -\left[\sum_{z\neq y} R_t(z, y) \cdot \frac{p_t(\zeta|y)^\alpha}{p_t(\zeta|z)^\alpha} \cdot g_t(z) - Q_t(y) g_t(y) + \left[Q_t(y) - R_t(y) - \alpha \partial_t \ln p_t(\zeta | y)\right] g_t(y)\right]. \\
\end{align}
Notice that $g$ reduces to a system of ODEs and thus, by the Picard-Lindel\"{o}f theorem, has a unique solution. Recall that $\gZ_1^\alpha = \sum_x p_1(x) p_1(\zeta|x)^\alpha = p(\zeta)^\alpha$, writing $p(\zeta)$ as the density of $\zeta$. Substituting $g_t(y) = p_t(y) \cdot p_t(\zeta|y)^\alpha / \gZ_1^\alpha$, then,
\begin{align}
    \text{L.H.S} &= \partial_t g(y) \\
    &= \frac{1}{\gZ_1^\alpha}\left[p_t(\zeta|y)^\alpha \partial_t p_t(y) + p_t(y) \partial_t ( p_t(\zeta|y)^\alpha)\right] \\ 
    \text{R.H.S} &= -\frac{1}{\gZ_1^\alpha}\left[\sum_{z\neq y} R_t(z, y) \cdot \frac{p_t(\zeta|y)^\alpha}{p_t(\zeta|z)^\alpha} \cdot g_t(z) - Q_t(y) g_t(y) + \left[Q_t(y) - R_t(y) - \alpha \partial_t \ln p_t(\zeta | y)\right] g_t(y)\right] \\
    &= -\frac{1}{\gZ_1^\alpha}\left[p_t(\zeta|y)^\alpha\sum_{z\neq y} R_t(z, y) p_t(z) - p_t(\zeta|y)^\alpha\sum_{z \neq y} R_t(y, z) p_t(y) - \alpha \partial_t \ln p_t(\zeta|y) p_t(y) p_t(\zeta|y)^\alpha\right] \\ 
    &= \frac{1}{\gZ_1^\alpha}\left[p_t(\zeta|y)^\alpha \partial_t p_t(y) + p_t(y) \partial_t ( p_t(\zeta|y)^\alpha)\right] \\ 
    &= \text{L.H.S}
\end{align}
Since  $p_1(y) = p_1(y \mid \zeta)$, it follows that the boundary condition $g_1(y) = p_1(y)$ is fulfilled. Then,
\begin{align}
\E[e^{A_t}] = \sum_x \int_\R e^a f_t(x, a) \dd a = \sum_x g_t(x) = \sum_x p_t(x) p_t(\zeta|x)^\alpha / \gZ_1^\alpha = \gZ_t^\alpha / \gZ_1^\alpha
\end{align}
Similarly, for an arbitrary function $h$ if we define $g^h_t(x) = \int_\R e^a h(x) f_t(x, a)\dd a$, we may show that
\begin{align}
\E[e^{A_t}h(x)] = \sum_x \int_\R e^a h(x) f_t(x, a) \dd a = \sum_x h(x) p_t(x) p_t(\zeta|x)^\alpha / \gZ_1^\alpha. 
\end{align}
Combining the two equations, we have,
\begin{align}
\frac{\E[e^{A_t}h(x)]}{\E[e^{A_t}]} = \frac{1}{\gZ_t^\alpha} \sum_x h(x) p_t(x) p_t(\zeta|x)^\alpha,
\end{align}
as desired.

To finish off the proof, we show that following:
\begin{enumerate}
    \item $\forall x \neq y. \; \alpha(\ln p_t(\zeta|y) - \ln p_t(\zeta|x)) = \alpha (\ln R^{\alpha=1}_t(x, y) - \ln R_t(x, y))$
    \item $\partial_t \ln p_t(\zeta | x_t) = \alpha R^{\alpha=1}_t(x)- \alpha R_t(x)$
\end{enumerate}
The first equation follows from the definition of the guided ratio matrix $R^{\alpha=1}_t(x)$. To show that $\partial_t \ln p_t(\zeta | x_t) = \alpha R_t(x) - \alpha R^{\alpha=1}_t(x)$, recall,
\begin{align}
    p_t(\zeta | x_t) = \E[p_t(\zeta|X_0)|X_t=x_t],
\end{align}
Therefore,
\begin{align}
\partial_t \ln p_t(\zeta|x_t) &= \frac{1}{p_t(\zeta|x_t)} \partial_t p_t(\zeta|x_t) \\ 
&= \frac{1}{p_t(\zeta|x_t)} \sum_{y} R_t(x_t, y) p_t(\zeta | y) && \text{(Reverse-Time) Kolmogorov Backward Equation} \\ 
&= R^{\alpha=1}_t(x_t) -  R_t(x_t) && \text{Definition of } R_t^{\alpha=1} 
\end{align}
This concludes the proof.

\section{Statement and Derivation of Discretised Weights} \label{sec:discretised-weights}
In this appendix, we show a discrete time version of \Cref{prop:cont-resampling} and the proof.

\begin{proposition} \label{prop:discrete-resampling}
Define $\{\hat{W}_{t_k}\}_{k \in \{1, \dots, T\}}$ as:
\begin{align*}
    \hat{W}_0 &= 1, \\ 
    \hat{W}_{t_k} &= \prod_{1 \leq j < k} \frac{p_{t_{j+1}|t_j}(Y_{t_{j+1}} | Y_{t_j})}{q_{t_{j+1}|t_j}(Y_{t_{j+1}} | Y_{t_j})} \notag \\ 
    &\quad \cdot  \prod_{1 \leq j < k}  \left[\frac{p_{t_{j+1}|t_j}(Y_{t_{j+1}} | Y_{t_j}, \zeta)}{p_{t_{j+1}|t_j}(Y_{t_{j+1}} | Y_{t_j})}\right]^\alpha.
\end{align*}
If $p_1(x_1) = p_1(x_1 | \zeta)$, then for any function $h : \mathcal{X} \to \mathbb{R}$ and $k \in \{1, \dots, T\}$,
\begin{align*}
\frac{1}{\gZ_{t_k}^\alpha} \sum_x h(x) p_{t_k}(x) p_{t_k}(\zeta | x)^\alpha = \frac{\mathbb{E}\left[\hat{W}_{t_k} h(Y_{t_k})\right]}{\mathbb{E}\left[\hat{W}_{t_k}\right]},
\end{align*}
where the expectation is taken over the law of $(Y_t, \hat{W}_t)$.
\end{proposition}
\begin{proof}
First note that,
\begin{align}
p_{t_{j+1}|t_j}(Y_{t_{j+1}} | Y_{t_j}, \zeta) &= p^\text{corrupt}_{t_j|t_{j+1}}(Y_{t_j} | Y_{t_{j+1}})\frac{M_t[p_0(x_0|\zeta)](Y_{t_{j+1}})}{M_t[p_0(x_0|\zeta)](Y_{t_j})} \\
&= p^\text{corrupt}_{t_j|t_{j+1}}(Y_{t_j} | Y_{t_{j+1}}) \frac{p_t(Y_{t_{j+1}}) p_t(\zeta|Y_{t_{j+1}})}{p_t(Y_{t_{j}}) p_t(\zeta|Y_{t_{j}})} \\ 
&= p_{t_{j+1}|t_j}(Y_{t_{j+1}} | Y_{t_j}) \frac{p_t(\zeta|Y_{t_{j+1}})}{p_t(\zeta|Y_{t_{j}})}.
\end{align}
Then it follows that,
\begin{align}
\hat{W}_{t_k} &= \prod_{1 \leq j < k} \frac{p_{t_{j+1}|t_j}(Y_{t_{j+1}} | Y_{t_j})}{q_{t_{j+1}|t_j}(Y_{t_{j+1}} | Y_{t_j})}\prod_{1 \leq j < k} \left[\frac{p_t(\zeta|Y_{t_{j+1}})}{p_t(\zeta|Y_{t_{j}})}\right]^\alpha \\
&= \prod_{1 \leq j < k} \frac{p_{t_{j+1}|t_j}(Y_{t_{j+1}} | Y_{t_j})}{q_{t_{j+1}|t_j}(Y_{t_{j+1}} | Y_{t_j})} \left[\frac{p_{t_k}(\zeta | Y_{t_k})}{p_1(\zeta | Y_1)}\right] ^\alpha \\ 
&= \prod_{1 \leq j < k} \frac{p_{t_{j+1}|t_j}(Y_{t_{j+1}} | Y_{t_j})}{q_{t_{j+1}|t_j}(Y_{t_{j+1}} | Y_{t_j})} \left[\frac{p_{t_k}(\zeta | Y_{t_k})}{p(\zeta)}\right] ^\alpha  && Y_1 = X_1`\perp \zeta \\ 
&=  \frac{1}{\gZ_1^\alpha}\prod_{1 \leq j < k} \frac{p_{t_{j+1}|t_j}(Y_{t_{j+1}} | Y_{t_j})}{q_{t_{j+1}|t_j}(Y_{t_{j+1}} | Y_{t_j})} && \gZ_1^\alpha \left[p_{t_k}(\zeta | Y_{t_k})\right]^\alpha = \sum_x p_1(x) p_1(\zeta|x)^\alpha = p(\zeta)^\alpha.
\end{align}
% \fv{$\gZ_1^\alpha = \sum_x p_1(x) p_1(\zeta|x)^\alpha$, and $p(\zeta)^\alpha= \left(\sum_x p_1(x) p_1(\zeta|x)\right)^\alpha$  , $\gZ_1^\alpha \neq p(\zeta)^\alpha$ }
Then,
\begin{align*}
\mathbb{E}\left[{\hat{W}_{t_k}} h(Y_{t_k})\right] &= \sum_{Y_{t_k}} {\hat{W}_{t_k}} h(Y_{t_k}) q_{t_k}(Y_{t_k})  \\
&= \frac{1}{\gZ_1^\alpha}\sum_{\{y_{i}\}_{i=1}^{k}:y_k=Y_{t_k}} \hat{W}_{t_k} h(Y_{t_k}) \prod_{j=1}^{k-1} q_1(y_1)q_{t_{j+1}|t_j}(y_{t_{j+1}} | y_{t_j}) \\ 
&= \frac{1}{\gZ_1^\alpha}\sum_{\{x_{i}\}_{i=1}^{k}:x_k=Y_{t_k}} h(Y_{t_k}) p_t(\zeta|Y_{t_k})^\alpha q_1(x_1) \prod_{j=1}^{k-1} p_{t_{j+1}|t_j}(x_{t_{j+1}} | x_{t_j}) \\
&= \frac{1}{\gZ_1^\alpha}\sum_{Y_{t_k}} h(Y_{t_k}) p_t(\zeta|Y_{t_k})^\alpha p_t(Y_k) && \text{Both }Y_t\text{ and }X_t\text{ starts at } p_1 \\ 
\end{align*}
Similarly,
\begin{align}
\mathbb{E}\left[\hat{W}_{t_k} h(Y_{t_k})\right] = \gZ_t^\alpha / \gZ_1^\alpha.
\end{align}
Combining the two equation lends us to,
\begin{align}
    \frac{1}{\gZ^\alpha_{t_k}} \sum_x h(x) p_{t_k}(x) p_{t_k}(\zeta | x)^\alpha = \frac{\mathbb{E}\left[\hat{W}_{t_k} h(Y_{t_k})\right]}{\mathbb{E}\left[\hat{W}_{t_k}\right]}
\end{align}
This concludes the proof.
\end{proof}

\newpage
\section{Sampling with Approximate Transition Discretised Weight} \label{sec:discretised-main-algorithm}

The exact transition probabilities $p_{t_{j+1}|t_j}$ and $q_{t_{j+1}|t_j}$ is typically unknown. In practice, we approximate these transitions using numerical methods such as Euler sampling. As the time step $\deltat$ approaches zero, these approximations converge to the true weights $W_t$. In \Cref{alg:the-approximate-algorithm}, we show the pseudocode of \Cref{alg:the-algorithm} using the approximated transition probability $\tilde{q}_{t|s}$ and $\tilde{p}_{t  |  s}$.

\begin{algorithm}[h!]
  \caption{Main Algorithm with Approximate Transition}
  \label{alg:the-approximate-algorithm}
  \begin{algorithmic}[1]
    \REQUIRE Number of particles $K$; Approximate Proposal Transition $\tilde{q}_{t|s}$; Approximate transition $\tilde{p}_{t  |  s}$; Temperature $\alpha$; Time-grid $\{t_l\}_{l=1}^T$; ESS threshold $\text{ESS\_THRESHOLD}$; Resampling algorithm \texttt{resample}.
    \STATE \textbf{Initialisation:}
    \STATE Sample $\{x_1^{(i)}\}_{i=1}^K$ i.i.d. from $p_1$.
    \STATE Set $\hat{w}_1^{(i)} \leftarrow 0$ for $i=1,\dots,K$.
    \FOR{$l = 1$ to $T$}
      \FOR{$i = 1$ to $K$}
        \STATE \textcolor{gray!80}{\COMMENT{Step 1: Propagate}}
         \STATE Sample $x_{l+1}^{(i)} \sim \tilde{q}_{t_{l+1} | t_l}\!\bigl(\cdot  |  x_l^{(i)}\bigr)$ 
        \STATE \textcolor{gray!80}{\COMMENT{Step 2: Update and Normalize Weights}}
        \STATE Set $\hat{w}_{l+1}^{(i)} \leftarrow \hat{w}_{l}^{(i)}  \cdot \frac{\tilde{p}_{t_{l+1} | t_l}({x^{(i)}_{l+1}} | {x^{(i)}_{l}})}{\tilde{q}_{t_{l+1}|t_l}\!\bigl({x^{(i)}_{l+1}}  |  x_{l}^{(i)})} \cdot \left[  \frac{\tilde{p}_{t_{l+1} | t_l}({x^{(i)}_{l+1}} | {x^{(i)}_{l}}, \zeta)}{\tilde{p}_{t_{l+1}|t_l}\!\bigl({x^{(i)}_{l+1}} |  x_{l}^{(i)})}\right]^\alpha$
      \ENDFOR
      \STATE \textcolor{gray!80}{\COMMENT{Step 3: Resample; see \Cref{sec:resample} for other resampling algorithms}} 
      \STATE Set $\text{ESS} \leftarrow \left(\sum_{i=1}^K w^{(i)}_{l}\right)^2 / \sum_{i=1}^K (w^{(i)}_{l})^2$
      \IF{$\text{ESS} \leq \text{ESS\_THRESHOLD}$}
          \STATE $\tilde{w}_{l}^{(i)} \leftarrow \hat{w}_{l}^{(i)} / \sum_i \hat{w}_{l}^{(i)}$
          \STATE Set $x_{l+1}^{(i)}, \hat{w}_{{l+1}}^{(i)} \leftarrow \texttt{resample}(\{x_{l}^{(i)}\}, \{\tilde{w}_{l}^{(i)}\})$ 
      \ENDIF
    \ENDFOR
    \STATE \textbf{Output:} 
    \STATE Particles $\{x_T^{(i)}\}_{i=1}^K$
  \end{algorithmic}
  \label{algo:main-algo}
\end{algorithm}

\newpage

\section{Pseudocode for Resampling Algorithms} \label{sec:resample}
In this section, we introduce various resampling strategies for Sequential Monte Carlo methods. Ultimately, the goal of resampling algorithm is to replace low weighted particles with high weighted particles to reduce the variance of weights such that the expectation of any function $\phi$ remains unchanged.

More specifically, consider at iteration $l$ of a SMC algorithm, given $K$ particles $\{x^{(i)}_l\}_{i=1}^K$ and $\{w^{(i)}_l\}_{i=1}^K$ from some joint distribution $\mathbb{Q}$ induced by the SMC algorithm. A resampling algorithm returning $\{\tilde{x}^{(i)}\}_{i=1}^K$ and $\{\tilde{w}^{(i)}\}_{i=1}^K$ generates consistent samples as long as for any bounded and continuous function $\phi$, $\E^\mathbb{Q}[w^{(i)}\phi(x^{(i)})] = \E^\mathbb{Q}[\tilde{w}^{(i)}\phi(\tilde{x}^{(i)})]$ where $\E^\mathbb{Q}$ notes an expectation on $\mathbb{Q}$.

\subsection{Multinomial Resampling}
The simplest and most common resampling strategy is the multinomial resampling, which involves the particles in the next iteration independently following a categorical distributions distributed according to the normalised weights.

\begin{algorithm}[h!]
  \caption{Multinomial Resampling}
  \label{alg:multinomial}
  \begin{algorithmic}[1]
    \REQUIRE Number of Particles $K$; Particles $\{x^{(i)}\}_{i=1}^K$; Normalised Weights $\{w^{(i)}\}_{i=1}^K$ 
    \FOR{$i = 1$ to $K$}
        \STATE Sample $\tilde{x}^{(i)} \sim \sum_{j=1}^K w^{(i)} \delta_{x^{(i)}} (\dd x)$
        \STATE $\tilde{w}^{(i)} = 1/K$
    \ENDFOR
    \STATE \textbf{Output:} 
    \STATE Particles $\{\tilde{x}^{(i)}\}_{i=1}^K$ and Weights $\{\tilde{w}^{(i)}\}_{i=1}^K$
  \end{algorithmic}
\end{algorithm}

\subsection{Partial Resampling}
We further consider employing a partial resampling strategy adapted from \citep{partal-resampling}, which we found to prevent mode collapse of samples effectively.
\begin{algorithm}[h!]
  \caption{Partial Resampling}
  \label{alg:partial}
  \begin{algorithmic}[1]
    \REQUIRE Number of Particles $K$; Particles $\{x^{(i)}\}_{i=1}^K$; Normalised Weights $\{w^{(i)}\}_{i=1}^K$; Resample Size $M$
    \STATE Set Resample Indices $I = \{i  |  w^{(i)} \text{ among the } \lfloor{M/2}\rfloor \text{ highest or } \lceil M/2 \rceil \text{ lowest} \}$
    \FOR{$i = 1$ to $K$}
        \IF{$i \in I$}
            \STATE Sample $\tilde{x}^{(i)} \sim \sum_{j=1}^K w^{(i)} \delta_{x^{(i)}} (\dd x)$
            \STATE Set $\tilde{w}^{(i)} \leftarrow  1/M \sum_{i\in I} w^{(i)}$
        \ELSE
            \STATE Set $\tilde{x}^{(i)} \leftarrow x^{(i)}$
            \STATE Set $\tilde{w}^{(i)} \leftarrow w^{(i)}$
        \ENDIF
    \ENDFOR
    \STATE \textbf{Output:} 
    \STATE Particles $\{\tilde{x}^{(i)}\}_{i=1}^K$ and Weights $\{\tilde{w}^{(i)}\}_{i=1}^K$
  \end{algorithmic}
\label{algo:partial_resampling}
\end{algorithm}
% \newpage
% \section{Additional MNIST generation result}
% \begin{figure*}[h!]
%     \centering
%     \includegraphics[width=\linewidth]{additional-mnist-generation-result.png}
%     \caption{\textbf{Accuracy on Class-Conditional Image generation}}
%     \label{fig:additional-mnist-generation}
% \end{figure*}

\newpage
\section{Additional text generation result}
\label{app:deft_result}

We display results on text generation control when using the DEFT guidance term. We find that:

For toxicity-controlled generation, SMC-based methods outperform guidance on control metrics across the guidance temperature range $[0.0, 2.0]$, which is even more apparent for higher SMC temperature $\alpha$. This is at the expense of a slightly degraded perplexity compared to guidance. At high guidance temperature, the guidance method is on par with SMC-based methods on toxicity score, at the expense of a significantly deteriorated perplexity. This indicates that at high guidance temperatures, the samples deviate from the fine-tuned data distribution. In that scenario, a good balance between perplexity and control generation is in the mid-to-low range of guidance temperature.

\begin{figure}[h!]
    \centering
    \includegraphics[width=\linewidth]{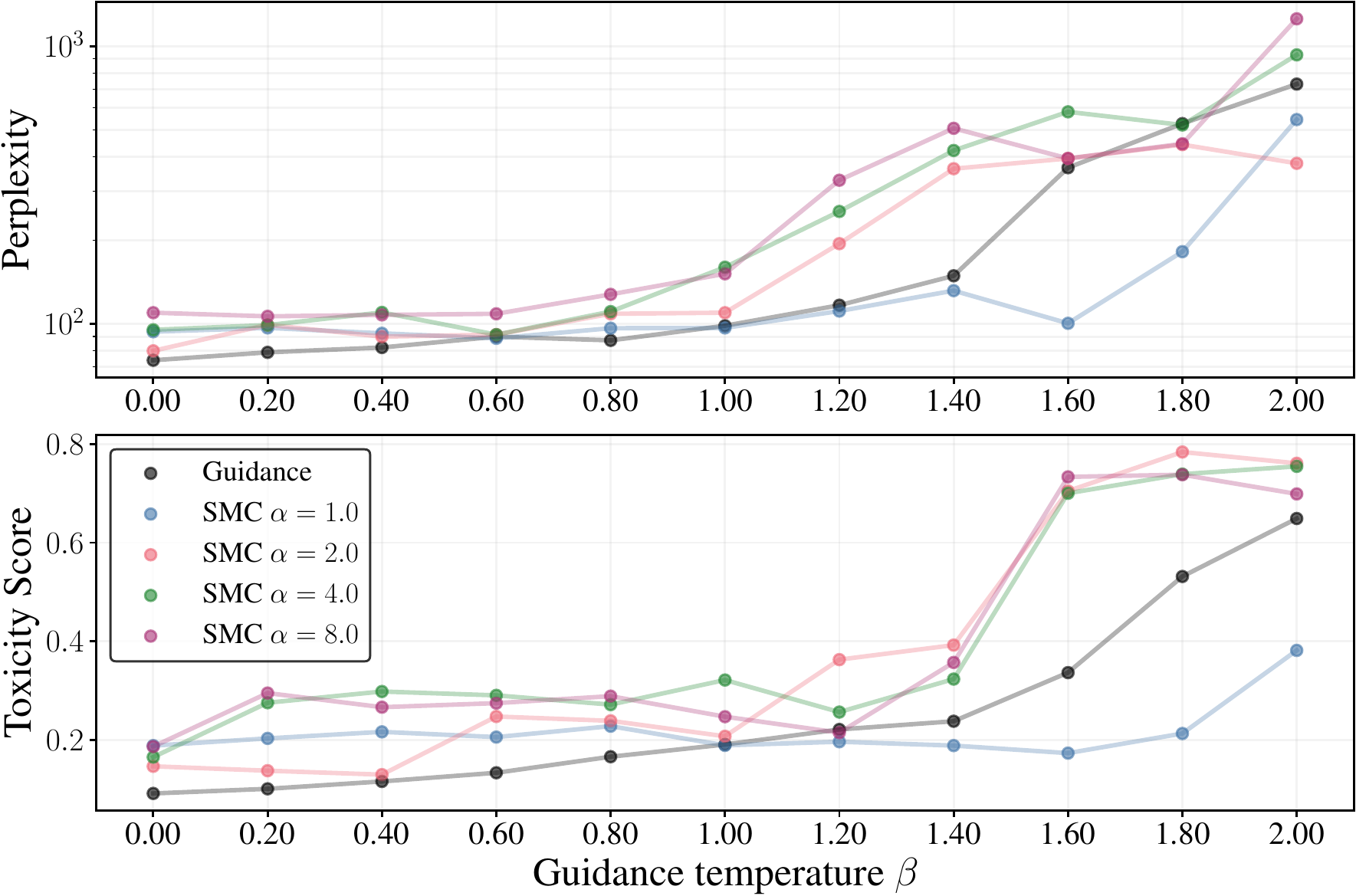}
    \vspace{-1em}
    \caption{\textbf{Toxicity controlled generation:} Guidance is performed with DEFT. SMC methods show improved toxicity scores across all proposal temperatures $\beta$, particularly for high $\beta$, demonstrating enhanced generation control.}
    \label{fig:toxic-deft}
\end{figure}

For text infilling, SMC-based methods achieve superior accuracy for guidance temperatures  $\beta \in [0.2, 0.8]$. However, at high guidance temperatures, both SMC and guidance methods attain comparable optimal accuracy and perplexity. In this specific scenario, despite lacking theoretical guarantees, the guidance method proves more practical due to its lower computational cost while maintaining equivalent performance.

\begin{figure}[h!]
    \centering
    \includegraphics[width=1\linewidth]{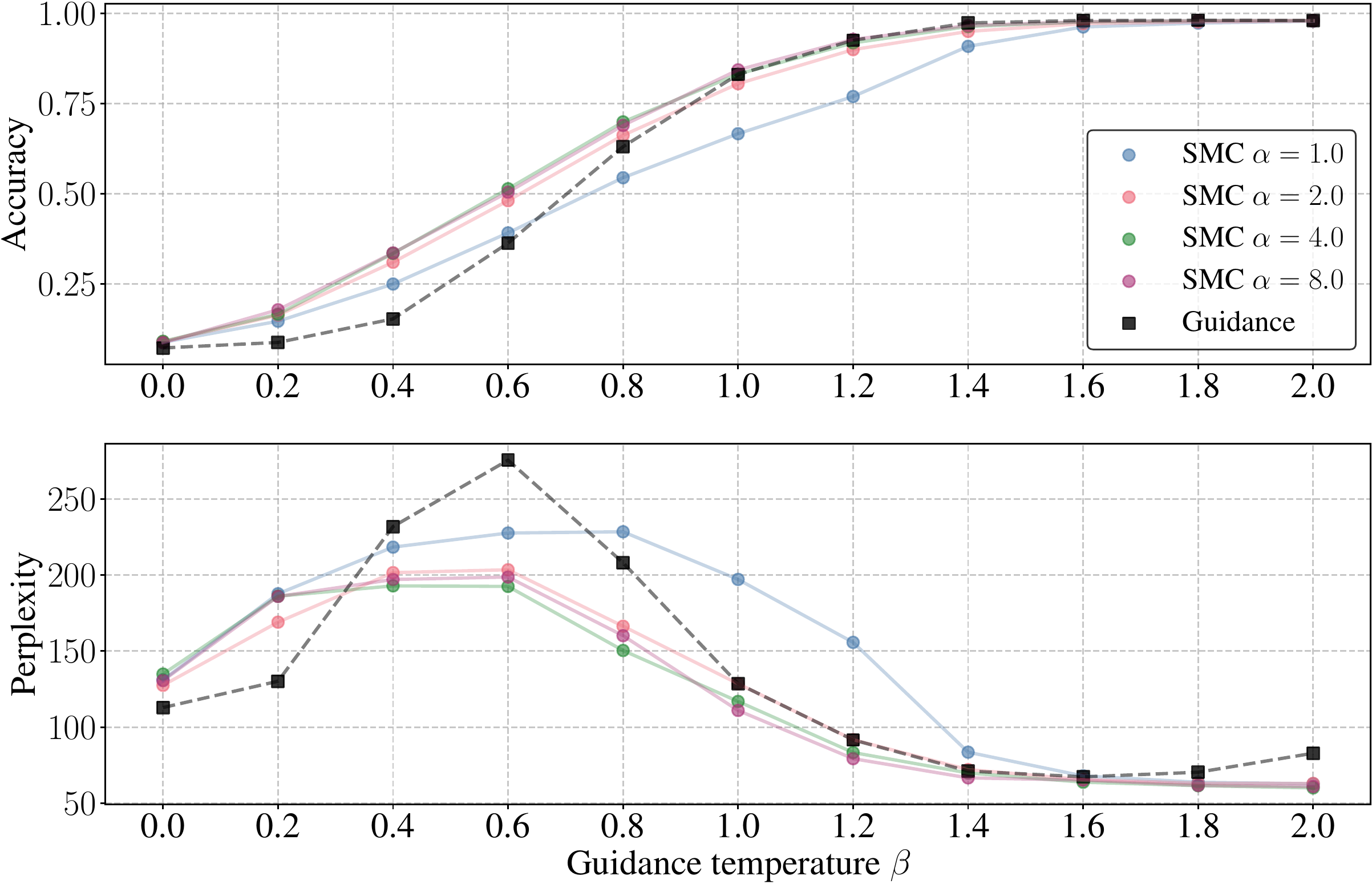}
     \vspace{-1.5em} 
    \caption{\textbf{Text infilling:} Guidance is performed with DEFT. SMC based methods show improved accuracy for $\beta \in \{0.2, 0.4, 0.6, 0.8\}$ while keeping a lower perplexity for $\beta \in  \{0.4, 0.6\}$ }
    \label{fig:deft-infill}
\end{figure}

For sentiment-controlled generation, SMC methods achieve higher accuracy for guidance temperatures $\beta < 0.6$, which comes at the cost of a slight increase in perplexity. At higher guidance temperatures ($\beta > 1.2$), both SMC and standard guidance converge to similar accuracy levels around 0.85, with comparable perplexity until $\beta = 1.6$. Beyond this point, all methods show significant perplexity deterioration, with SMC methods exhibiting slightly higher perplexity than standard guidance.

\begin{figure}[h!]
    \centering
    \includegraphics[width=\linewidth]{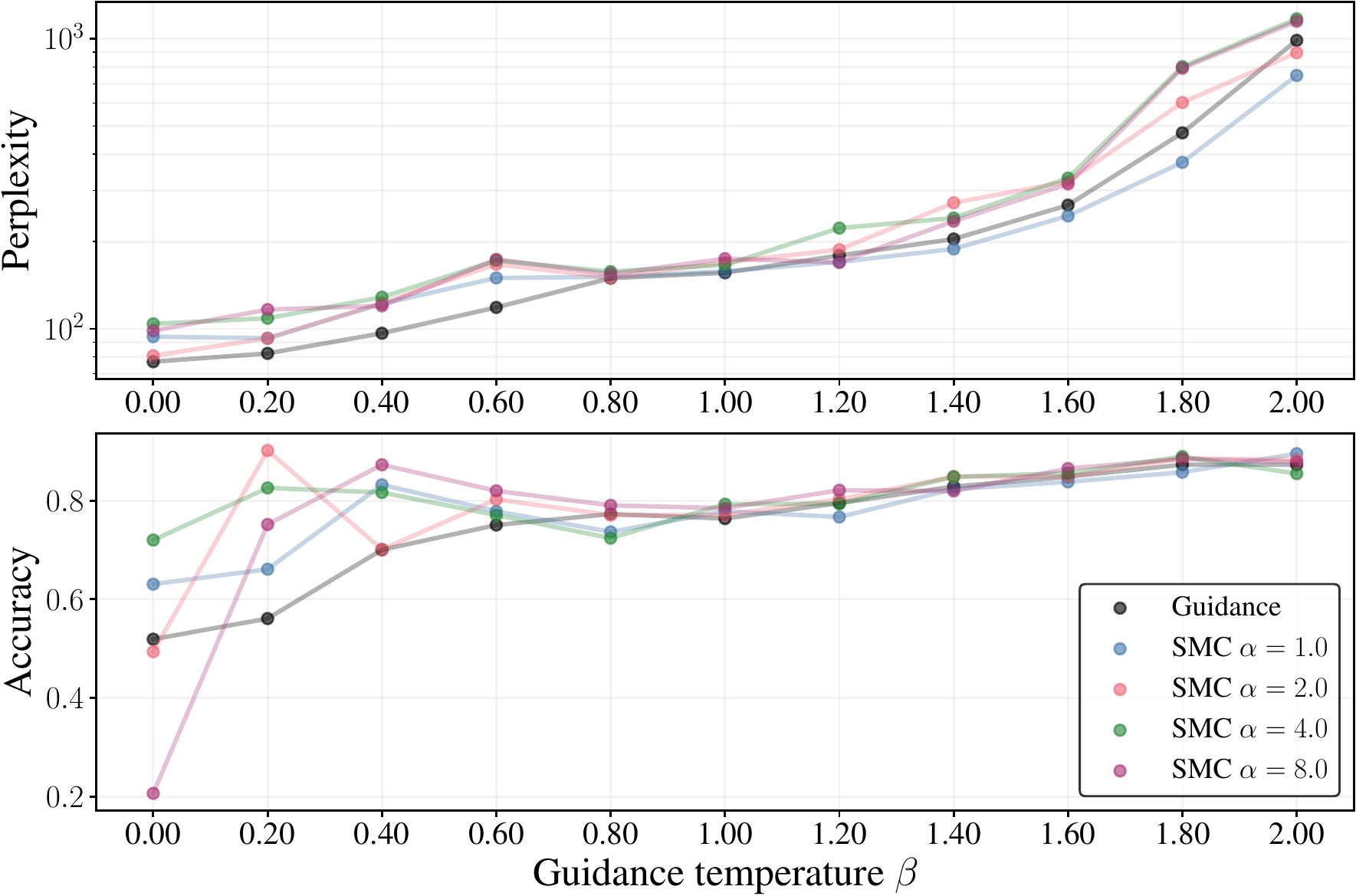}
    \vspace{-1em}
    \caption{\textbf{Sentiment controlled generation}: Guidance is performed with DEFT. For guidance temperature lower than $0.6$ SMC methods show improved accuracy while having higher perplexity. For larger guidance temperature accuracy and perplexity are relatively similar.}
        \label{fig:sentiment-deft}
\end{figure}

\newpage\phantom{blabla}
\newpage\phantom{blabla}

\section{Samples}
\begin{figure}[h]
    ant \texttt{EOT} a very funny and rewarding small comedy . \texttt{EOT} a rather else altogether \texttt{EOT} genuinely sweet and \texttt{EOT} evokes its low-budget constraints fairly well \texttt{EOT} , evocative \texttt{EOT} 's spooky . \texttt{EOT} an interesting concept \texttt{EOT} here 's the treat . \texttt{EOT} a solidly entertaining film in its own right \texttt{EOT} supremely goofy pleasure \texttt{EOT} demonstrated extraordinary faith \texttt{EOT} an unflinching meditation on how business can change those who enter \texttt{EOT} love the film . \texttt{EOT}  -- deep down you 've a family \texttt{EOT} a fascinating glimpse \texttt{EOT} of place with its oscar-worthy predecessors \texttt{EOT} pounding, offering fine acting moments , \texttt{EOT} it 's a surest bet \texttt{EOT} an engaging fantasy \texttt{EOT} , the material actually springs to life \texttt{EOT} director credit gyllen brings together caine and grant \texttt{EOT} star charisma \texttt{EOT} consistent \texttt{EOT} what 's best about dustin gondry 's sophisticated read my lips \texttt{EOT}  \texttt{EOT}  ... scotches . \texttt{EOT} 's surprisingly faithful \texttt{EOT}  \texttt{EOT} dazzling \texttt{EOT} make up for the execution 's weak comic buttons . \texttt{EOT} curiously funny, intriguing , \texttt{EOT} just that satisfying exploration of love and companionship \texttt{EOT}  \texttt{EOT} emotional sympathy \texttt{EOT} make it entertaining , \texttt{EOT} ungainly work \texttt{EOT} strongest films \texttt{EOT} funny \texttt{EOT} to sharp writing \texttt{EOT} make for a great director 's to make, everlasting \texttt{EOT} innocence \texttt{EOT} though not technically proficient or as \texttt{EOT} remark to ` believer ' in some time \texttt{EOT} sometimes insightful writer \texttt{EOT} mr. honorably \texttt{EOT} and unforgettable characters \texttt{EOT} cool, cocky, and -- dare i say twice -- hollywood-esque \texttt{EOT} powerful story \texttt{EOT} surprisingly sweet \texttt{EOT} engaging, compelling tale \texttt{EOT} beautiful \texttt{EOT} vibrant air \texttt{EOT} but ultimately winning film \texttt{EOT} its simple expressiveness \texttt{EOT} enriched than limited by its exceptional lead performances \texttt{EOT} of impressively challenging social drama \texttt{EOT} the rich formalism \texttt{EOT}  \texttt{EOT} keep you \texttt{EOT} of laughs \texttt{EOT} quieter \texttt{EOT} the praises \texttt{EOT} some monster movie \texttt{EOT} as his penchant for documentaries \texttt{EOT} is latin star to young international hip-hop audiences \texttt{EOT} often funny romantic comedy \texttt{EOT} smoother \texttt{EOT} only enhance his good looks \texttt{EOT} immensely enjoyable \texttt{EOT} one of the most idiosyncratic comedies \texttt{EOT} oddly watchable \texttt{EOT} ambitious \texttt{EOT} directed personalities \texttt{EOT} interesting, i must say , \texttt{EOT} faithful , \texttt{EOT} well, \texttt{EOT} well lived \texttt{EOT} compelling and gripping \texttt{EOT} everyone 's comic relief \texttt{EOT} truly funny \texttt{EOT} enjoy \texttt{EOT} elegant delivery \texttt{EOT} idea \texttt{EOT} familiarity \texttt{EOT} to hold you well past its 90 minutes \texttt{EOT} find love \texttt{EOT} fresh sense \texttt{EOT} has spliceed together bits and pieces of hagney 's reign \texttt{EOT} for good, scary movies \texttt{EOT} gaining this movie \texttt{EOT} flaccid satire , \texttt{EOT}  \texttt{EOT} respectable, sweetly adventurous \texttt{EOT} a film \texttt{EOT} smart and subtle \texttt{EOT} vibrant and \texttt{EOT} their fathers \texttt{EOT} her maternal fury of faith \texttt{EOT}  \texttt{EOT} protagonists \texttt{EOT} is surprisingly insightful and fun and entertaining \texttt{EOT} that acts as if it never had any idea what \texttt{EOT} , originality and technical skill \texttt{EOT} argue \texttt{EOT} above all, blue crush is nothing to overlook . \texttt{EOT} sulky and beautiful rendering of an intense mystery about the world 's greatest teacher \texttt{EOT} daring teacher \texttt{EOT} cinematic stamp \texttt{EOT} warm , \texttt{EOT} its inviting \texttt{EOT} sexy \texttt{EOT} is realistically terrifying \texttt{EOT} a surprising entree \texttt{EOT} invigoratingly irresistible \texttt{EOT} action-power appeal \texttt{EOT} slyly debated film \texttt{EOT} mention, engaging premise \texttt{EOT} , there are enough laughs to keep on clicking through \texttt{EOT} puts us off-center in the unfolding of bielinsky 's clever scenario , \texttt{EOT} the local flavor and french musicmaking traditions \texttt{EOT} elegance \texttt{EOT} it has a real `` story '' \texttt{EOT} at least the funniest idea \texttt{EOT} works well on different levels, until it is actually funny . \texttt{EOT} , cleverly crafted and \texttt{EOT} scenes brimmed \texttt{EOT} allows you to forgive its basic humanity \texttt{EOT} that tickles the chills of \texttt{EOT} subtle \texttt{EOT} the film a stunning lesson in human-scale acting , \texttt{EOT} 's a film school exercise with dignity and wit that `` should have been right to grow up \texttt{EOT} a riot like john ritter 's brothers ... \texttt{EOT} thumbs up \texttt{EOT} seemingly \texttt{EOT} they 're the target audience because \texttt{EOT} sensitive performances and \texttt{EOT} magnificent landscape \texttt{EOT} caric ballot \texttt{EOT} bring more goodies . \texttt{EOT} as cohesive , \texttt{EOT} some first-time director \texttt{EOT} of the finest kind \texttt{EOT} a spirited \texttt{EOT} 're out tonight \texttt{EOT} style and color \texttt{EOT} a cleverly written and thoroughly winning portrayal of titular kid a woodman who 's become a self image with kids \texttt{EOT} emotional evolution \texttt{EOT} she 's a tempting bouquet -- if you live a minute ... come and see the charm of the movie \texttt{EOT} is credible \texttt{EOT} a classical actress \texttt{EOT} delicious camera work \texttt{EOT} a treat , \texttt{EOT} a more absorbing piece \texttt{EOT} all builds up to the easy, endearing knockout . \texttt{EOT} still charming -- likely his next film wo n't be the greatest date movie \texttt{EOT} , sweet \texttt{EOT}  . \texttt{EOT} most good-hearted \texttt{EOT} a few good ideas \texttt{EOT} a feel-good thriller \texttt{EOT} , illuminating study \texttt{EOT} large metaphor \texttt{EOT} successful \texttt{EOT} fresh is `` fresh '' again \texttt{EOT} tell this tale \texttt{EOT} felt and affecting \texttt{EOT} 
    \caption{Sentiment control - positive samples - SMC $\beta = 1.0$, $\alpha = 2.0$. \texttt{EOT} indicates End Of Text}
\end{figure}

\begin{figure}[h]
     having a difficult time telling what time in the past year . \texttt{EOT} a pale successor \texttt{EOT} an odd bit of an erratic \texttt{EOT} killing time \texttt{EOT} wobbly \texttt{EOT} mean-spirited \texttt{EOT} 's no energy \texttt{EOT} carries little insight on either end of the other \texttt{EOT} dangerous comedy \texttt{EOT} a slightly flat , \texttt{EOT} the little the seams \texttt{EOT}  \texttt{EOT} it drags \texttt{EOT} perverse \texttt{EOT} this bland \texttt{EOT} the cheapness \texttt{EOT} poorly \texttt{EOT} any teen movie starring slackers or \texttt{EOT} succeeded \texttt{EOT} brain strain \texttt{EOT} ill writing \texttt{EOT} may also not have been the grossest american bike movie of 2002 . \texttt{EOT} absurdities \texttt{EOT} it 's too simpleminded -- \texttt{EOT} has the idea been sealed in a jar \texttt{EOT} this obvious rip-off \texttt{EOT} racist \texttt{EOT} is wit, or even good acting , \texttt{EOT} more like satire than illuminating examination \texttt{EOT} despite snow dogs, leaves us cold . \texttt{EOT} the action is predictable with many subplots and \texttt{EOT} very insecure \texttt{EOT} 's still flibbled together its clichés . \texttt{EOT} bad combination \texttt{EOT} particularly suspenseful at times \texttt{EOT} deep technical blues vs blues \texttt{EOT} are so smeary, blurry and jagged \texttt{EOT} form 51 is n't really scary enough for `` midnight flick '', and \texttt{EOT} thought you or i were paying full price for this, and rent those instead \texttt{EOT} is a minor miracle \texttt{EOT} handbags in saving private ryan \texttt{EOT} like ` to suck ' \texttt{EOT} its clichéd scenes of war \texttt{EOT} poorly dubbed dialogue, and \texttt{EOT} 's hard to imagine a product more repellent than adam sandler 's 2002 \texttt{EOT}  \texttt{EOT} weighted a sci-fi story that 's bottom-heavy and tired \texttt{EOT} wrong things \texttt{EOT} stiflingly \texttt{EOT}  \texttt{EOT} endless exposition \texttt{EOT} simply overcooked \texttt{EOT} evasive \texttt{EOT} call an `` ambitious failure \texttt{EOT}  \texttt{EOT} even less ambitious failure \texttt{EOT} ridiculous and sloppy \texttt{EOT} no air conditioning \texttt{EOT} the ragged pacing \texttt{EOT} a loud, crassly predictable rehash \texttt{EOT} how lame it all is \texttt{EOT} ends world traveler 's lackluster title \texttt{EOT} have done with all the subtlety from earlier had taken in favor of more user-friendly computer tutorial \texttt{EOT}  \texttt{EOT} mind settles into a irksome parade of human frailty and death . \texttt{EOT} smug and exploitative \texttt{EOT} the whining \texttt{EOT} this latest comedy tangents of pratfalls, injuries and pranks \texttt{EOT} to anyone who suffered through halfway through david rifkin 's sweet home ` alabama \texttt{EOT} old pickup \texttt{EOT} should have been ordered off the screen \texttt{EOT} like the plague \texttt{EOT} neges toward fuzziness \texttt{EOT} its skewed acting , \texttt{EOT} an odd, painful and obnoxious attempt at a documentary \texttt{EOT} , repetitious and contrived \texttt{EOT} lacks a moment of considerable poignancy . \texttt{EOT} chaos and \texttt{EOT}  \texttt{EOT} cold, emotionally opaque \texttt{EOT} stands nowhere \texttt{EOT} frustrating \texttt{EOT} dull \texttt{EOT} suffered \texttt{EOT} characteristic quirks \texttt{EOT} half-step \texttt{EOT} , it 's hopeless . \texttt{EOT} there 's nothing new to be taken from ` fatal attraction ' 4ever \texttt{EOT} is painfully bad . \texttt{EOT} unfaithful hollywood fluff \texttt{EOT} have little interest in jez begley 's book \texttt{EOT} no foundation for disney \texttt{EOT}  \texttt{EOT} , herzog is the director 's logistically and \texttt{EOT} is it not hagiographic, though \texttt{EOT} need more than plot \texttt{EOT} rocks and holes \texttt{EOT} with lots of solips \texttt{EOT} ving \texttt{EOT} at times maddeningly repetitive . \texttt{EOT} rendered all round square edges, blurry images and murky and \texttt{EOT} 's cloying, manipulative \texttt{EOT} trial \texttt{EOT} on sub-zero animation \texttt{EOT} an annoying , \texttt{EOT} went to the restroom but want my money back . \texttt{EOT} water torture \texttt{EOT} pompous \texttt{EOT} rigid idea \texttt{EOT} insufferable \texttt{EOT} as boring and pedestrian \texttt{EOT} , heavy-handed \texttt{EOT} marginal insights \texttt{EOT} from lost ring to snatch on stage \texttt{EOT} hampered by a shabby script \texttt{EOT} gold is a real hollywood dog story \texttt{EOT} limpid and pretentious endeavor \texttt{EOT} is subtler than it might have been . \texttt{EOT} the film with relentlessly nasty situations \texttt{EOT} 's not handled well \texttt{EOT} wastes dialogue \texttt{EOT} sensible violence \texttt{EOT} of all the male junkie post-about obsessive relationships \texttt{EOT} is a testament to the author 's work \texttt{EOT} have had room for more creative action . \texttt{EOT} empty farce \texttt{EOT} liar zone \texttt{EOT} could call it tacky \texttt{EOT} painfully awful \texttt{EOT} forget about the fact that kennedy has nothing to offer up his act \texttt{EOT} the faintest hopefulness , \texttt{EOT}  \texttt{EOT} flat \texttt{EOT} , spiteful idiots \texttt{EOT} is superficial and \texttt{EOT} describe characters ' frustrations \texttt{EOT} end zone \texttt{EOT} familiar and thoroughly recycled plot . \texttt{EOT} tired \texttt{EOT} aimless direction, pathetic acting, heavy dialogue, tissue-thin characters \texttt{EOT}  \texttt{EOT} of inept filmmaking \texttt{EOT} last 15 years \texttt{EOT} the picture just \texttt{EOT} nothing except they ca n't -- anything, really, seems to matter very much \texttt{EOT} a noticeable amount of its own \texttt{EOT} neat \texttt{EOT} best elsewhere \texttt{EOT} waste ; \texttt{EOT} human behavior , \texttt{EOT} too fancy \texttt{EOT} derivancy \texttt{EOT} distinct comic direction \texttt{EOT} this m-out-l lesbian comedy \texttt{EOT} does n't give a complete picture . \texttt{EOT} shallow plot \texttt{EOT} could have any interest in silly fluff \texttt{EOT} , tedious , \texttt{EOT} 's not clear what this turkey is
    \caption{Sentiment control - negative samples - SMC $\beta = 1.0$, $\alpha = 2.0$. \texttt{EOT} indicates End Of Text}
\end{figure}

\begin{figure}[h]
     or it really costs that far to finish rail. \texttt{EOT} I always thought the part time NDP would be on the hook for this, but then they had to master the ndp's fiscal policies. \texttt{EOT} Fess up........ \texttt{EOT} He managed it brilliantly. \texttt{EOT} You are wrong in that regard. Harper suggested enough to leave the potential harm to children (and adults) to see. \texttt{EOT} Agreed. Further increasing the income of Quebec will likely lead to increased income tax charges for Ontario governments attempts to do same. \texttt{EOT} rumps.. the globe loses again.. its only readership is NK \texttt{EOT} once again putting alt lunatic whining over the real picture bud \texttt{EOT} Alaska hasn't cut the PF yet Walker over spent our PFD for the same 40 years. Walker has 3 more session to do some cuts cutting back PFD dividends not PFD raiding the new gas line. Enough Permanent fund cut already Walker and Governor Walker and Governor Walker. Time for legislature and real leadership in Alaska needing new taxes. \texttt{EOT} Yes Bravo Orange Pail! \texttt{EOT} Sorry son but don't look forward to Ducks game, it's always good to find out who has the better quarterback. \texttt{EOT} Trump is fleeing the Trump administration right now. \texttt{EOT} Sorry, you've undermined the whole point of your comment ... \texttt{EOT} How about everyone familiar with Dale Wright? \texttt{EOT} One less Hawaii Democrat. \texttt{EOT} Where's Cory Gardner. \texttt{EOT} Silly Senate. \texttt{EOT} Happy New Air Canada - what a touch of sweet irony. \texttt{EOT} Great comments! \texttt{EOT} Seems like your actions out in front of the world for all people to see. \texttt{EOT} there is no illegal world building \texttt{EOT} Trudeau ? \texttt{EOT} Not to fast.....tax Vancouver got it fair share. \texttt{EOT} LOL....... \texttt{EOT} Maybe our Senators Cory and Murkowski will vote for a repeal. \texttt{EOT} This is about criminal charges underway and emails continue to be the target. Mueller will investigate the names Hillary, let Mueller have his look and hopefully does find Hillary exists. \texttt{EOT} I'm sure snakes are the least human kind. \texttt{EOT} Great pleasant surprise. \texttt{EOT} I agree with you! \texttt{EOT} The problem is that guys like Meredith and Crapwell will just fade away. \texttt{EOT} Alceste three times is running as a republican. \texttt{EOT} Wrong case. Can't find a place right there. \texttt{EOT} http://JustPeace.org That's Fairbanks, Fairbanks \texttt{EOT} Yep. His opening rhinos ring louder every posting. \texttt{EOT} Not China nor NK. Neither have threatened NK. \texttt{EOT} Maybe they break the law along with other citizens who drive motorized vehicles. \texttt{EOT} What were you wearing? Wear the ladies locker? \texttt{EOT} Get a clue who play each string on the violin. \texttt{EOT} I have been discouraged about owning a home when rental prices have gone down significantly. It seems like there is only reason for gray market rentals any more. If I could find an area that had a small amount of rental units and never close it with no decision but rent, then I could buy in. \texttt{EOT} More Liberal tabloidism fleeced by paid leftist disinformation efforts. Good riddance! \texttt{EOT} Didn't happen to read your post either. \texttt{EOT} Wrong Old Spice.... \texttt{EOT} IRT Gary Crum, we don't need any new information when it comes to responding to the DP. \texttt{EOT} why? \texttt{EOT} Then why tax the PFD earnings when Anchorage borough and developers pay local property taxes with PFD earnings? \texttt{EOT} Agreed! Pay you to Netflix any network programming than you get in actual services you pay for. \texttt{EOT} You wanted Hillary is a lot different than wishing Hillary lost...... \texttt{EOT} Does every car require a plate handle or do democrats have ? \texttt{EOT} They have not given up anything re producing soft fishing. The dairy industry will not be on any side of the deal when NAFTA gets a bad looks up. First side to lose the deal is at risk. \texttt{EOT} Agreed. \texttt{EOT} I assume it isn't you Gary \texttt{EOT} Yeahhhh trump juice! \texttt{EOT} I remember long ago the Cons and the NDP were slamming each other behind the scenes. \texttt{EOT} Your tolerance must be getting really old. \texttt{EOT} Wouldn't say Amy hasn't heard here. \texttt{EOT} So Berkie and Caldwell belong in jail? \texttt{EOT} the big question is "which will they write it" \texttt{EOT} How was this deficit announced 6 months ago? \texttt{EOT} Very good article thanks for posting Vic. \texttt{EOT} Completely untrue. South America is producing the most efficient portion of the world's oil soon. BTW, if your planning on expanding pipelines, move ahead and avoid Alberta. \texttt{EOT} Wow. Can't even conceive of imagining how naïve trumpsters are. \texttt{EOT} Nah. It was a big one. \texttt{EOT} Prove it. Tax my dividend too. \texttt{EOT} Did't the letter writer say Lame County? \texttt{EOT} Yep! That's who you "democratic" voted into office. \texttt{EOT}  \texttt{EOT} Agreed! Harper! Has to go. \texttt{EOT} Sorry but not quite. The data you are referencing is one game show. \texttt{EOT} So Alaska needs our own modern rail system? We don't need a modern system of satellites like the majority of lights on the nation. \texttt{EOT} I find it hilarious that idiot trump does nothing to mitigate
    \caption{Toxicity control - non toxic samples - SMC $\beta = 1.2$, $\alpha = 4.0$. \texttt{EOT} indicates End Of Text}
\end{figure}

\begin{figure}[h]
     when commuters were most busy......... \texttt{EOT} Eh. Used that logic. \texttt{EOT} Better hope the coming recession never happens \texttt{EOT} Enough with your infomeration Bombing homes works better. \texttt{EOT} Just keep Trump in. It could be our destiny to get rid of all Obama's puppet masters \texttt{EOT} Here's my only thought regarding your comment: Keep up! \texttt{EOT} How did I need any further contribution to that? \texttt{EOT} Someone please refute my fact. \texttt{EOT} The phony Liberals who "thought" a love of country above party will have no spine. As always, the press will focus on facts and Republicans will relentlessly call out real Liberalism. \texttt{EOT} That's a fact. Comey's firing does have been the direct result of Comey's conduct acting inappropriately for US intelligence. Comey is not the person fired out of political motivation, but Trump was rejected as for the US Criminal Code but rather the federal judge has rejected him. \texttt{EOT} ": An extension for the appeal period or even an extension for each waiver signed may be all Kobachak’s congressional breaks." As I pointed out in a Civil Comments form on behalf of Kobachak in 2016 there was nothing further to see in Kobachak's case. He picked a fight and all is well with his formerly a non-friend and ally. \texttt{EOT} So? The author is anonymous? at least he doesn't have a secret. BTW, you seem to assume that Kobachak doesn't share your own psychology. Darn it. \texttt{EOT} I am a retired member of the Sierra Club, although I am also a member of the California chapter at https://fight.org/ and the Northeast chapter at https://www.sanfrust.org/ The health issue in Puerto Rico is significant and I think about it from a personal perspective. Trump is really sparking passions in his comments so will face criticism. Aside from that aways, he needs to understand what actually took place in Texas City and events that resulted in this disaster there. \texttt{EOT} Our government is only better than you. \texttt{EOT} Now that Trudeau has been given free rein in unilateral withdrawal from the U.S., then why would we subject ourselves to further a significant trade relationship that the Chinese and Russians absolutely want? It can only work here if we remove big historical barriers such as the Suzuki's and tackle these for our benefit. The world is entering a new era of mutual respect. \texttt{EOT} This is more of exactly what's expected. \texttt{EOT} That's wording that selectively is used. Only a few extreme views dominate the discussion. These are only three of 10. \texttt{EOT} More that some people do disagree with the process as they may have alternative theories. I also suggest you get educated and understand the current science and the reality of climate change. \texttt{EOT} Aloha, go year long in different neighborhoods like season after year. I see day in and day out the exact same homeless; no different penalty for visiting Eugene; same as anywhere else. \texttt{EOT} Thanks for the linking C of B page to Quebec. Donna can't cut it. \texttt{EOT} The crime is not unusual, per say. They are allowed their privacy, but selectively. No tunnels to break, but it is a crime. \texttt{EOT} The IPCC failing to release public information regarding the case runs contrary to the article's recommendation. Quite clearly his fate is not to be ended in death as was indicated by the way he was raised when confronted with simply not being born. \texttt{EOT} Again, are you not proud of your experience and reach? \texttt{EOT} I agree. This analysis does not take into account that the content of most of our English-language newspapers is particularly bias. To what appears to be happening in Canada a few days, but every once in a while a Canadian newspaper is going to highlight certain facts because of the importance a story that few would know. It has also been said that many newspapers insist on showing all news every week and not just the stories that are the most important. \texttt{EOT} If i can looked at things differently and see more from the top - it emerges I see less as 'completely out of line' as others feel I have looked more towards the 'top'. \texttt{EOT} We are all doing the best we do, sometimes it takes the next thing. \texttt{EOT} Having a minimum wage for the smaller employees in Alaska must be expensive because the businesses have to do their work at home thru government, education and the overall economy. My monthly cost of organic food is \$10.71 per hour plus. So how can I go through the trouble and earn that many per hour rather than having a loss on these goods? I paid lucky so I will have to stock up while in retirement, but that is still very expensive. Remember you not only need the help with government but with income, earned through education, and the rest of the economy. \texttt{EOT} Sure, a lot of kids are taking hitches and missing their focus. The selection of character, athleticism and tougher focus is what determines the chance of success. \texttt{EOT} It's not everything. He did hate the west in his first election. \texttt{EOT} It's the First Amendment folks to blame ... I say to "vide wolf, flock"
    \caption{Toxicity control - non toxic samples - Guidance $\beta = 1.2$. \texttt{EOT} indicates End Of Text}
\end{figure}

\end{document}